\documentclass[11pt]{article}

\usepackage[final]{acl}

\usepackage{amsmath,amsfonts,bm}

\def\eqref#1{equation~\ref{#1}}

\def\1{\bm{1}}

\DeclareMathAlphabet{\mathsfit}{\encodingdefault}{\sfdefault}{m}{sl}
\SetMathAlphabet{\mathsfit}{bold}{\encodingdefault}{\sfdefault}{bx}{n}

\usepackage{times}
\usepackage{latexsym}

\usepackage[T1]{fontenc}

\usepackage[utf8]{inputenc}

\usepackage{microtype}

\usepackage{inconsolata}

\usepackage{graphicx}

\usepackage{booktabs}
\usepackage{multirow}
\usepackage[table]{xcolor}

\usepackage{algorithm}
\usepackage{algpseudocode}
\usepackage{algorithmicx}

\usepackage{pifont}

\usepackage{subcaption}
\usepackage{amsfonts}
\usepackage{url}

\usepackage{soul}
\usepackage[most]{tcolorbox}
\tcbuselibrary{skins}
\usepackage{enumitem}

\title{LegalGraphRAG: Multi-Agent Graph Retrieval-Augmented Generation for Reliable Legal Reasoning}

\author{
  \textbf{Zerui Chen\textsuperscript{1}},
  \textbf{Qinggang Zhang\textsuperscript{3$\dagger$}},
  \textbf{Zhishang Xiang\textsuperscript{2}},
  \textbf{Zhimin Wei\textsuperscript{1}},
  \textbf{Linfeng Gao\textsuperscript{2}}, \\
  \textbf{Xiao Huang\textsuperscript{3}},
  \textbf{Zhihong Zhang\textsuperscript{1$\dagger$}},
  \textbf{Jinsong Su\textsuperscript{1$\dagger$}} \\
  \textsuperscript{1}School of Informatics, Xiamen University \\
  \textsuperscript{2}Institute of Artificial Intelligence, Xiamen University \\
  \textsuperscript{3}The Hong Kong Polytechnic University \quad
  \\
  \texttt{chenzerui1@stu.xmu.edu.cn}; \\
  \texttt{qinggang.zhang@polyu.edu.hk}; 
  \texttt{\{zhihong,jssu\}@xmu.edu.cn}
}

\newcommand\blfootnote[1]{%
  \begingroup
  \renewcommand\thefootnote{}\footnote{#1}%
  \addtocounter{footnote}{-1}%
  \endgroup
}

\definecolor{headerblue}{RGB}{220, 240, 250}
\definecolor{darkgreen}{RGB}{100, 200, 100}
\definecolor{midgreen}{RGB}{150, 230, 150}
\definecolor{lightgreen}{RGB}{200, 240, 200}
\definecolor{headerpurple}{RGB}{235, 230, 250}

\begin{document}
\maketitle
\blfootnote{$\dagger$ Corresponding author.}
\begin{abstract}


Graph-based Retrieval-Augmented Generation (GraphRAG) advances flat document retrieval by structuring knowledge as relational graphs, enabling more coherent and effective reasoning. However, applying it to specific domains like legal reasoning faces critical challenges. (i) Legal corpora are heterogeneous, containing multi-granular knowledge from cases, articles, and interpretations. A flat knowledge graph cannot adequately differentiate between factual details, applied rules, and abstract principles, limiting accurate retrieval. (ii) Reliable legal judgment demands transparent, evidence-based reasoning. Traditional RAG passes retrieved context directly to an LLM without verification, resulting in opaque, error-prone reasoning.
To this end, we propose \textbf{\texttt{LegalGraphRAG}}, a framework designed for reliable legal reasoning. Our approach introduces two core components: a hierarchical legal graph that hierarchically organizes legal sources to enable retrieval at appropriate abstraction levels, and a multi-agent system for reliable legal reasoning, where a Researcher retrieves candidate evidence, an Auditor rigorously verifies its validity against source documents, and an Adjudicator synthesizes the set of verified evidence to render a final judgment. Extensive experiments show that LegalGraphRAG achieves the state-of-the-art performance, outperforming existing GraphRAG baselines in accurate and trustworthy legal analysis. Our code, datasets and implementation details are available at \textcolor{blue}{\url{https://github.com/XMUDeepLIT/LegalGraphRAG}}.
\end{abstract}

\section{Introduction}
\begin{figure}[t]
    \centering
    \includegraphics[width=\linewidth]{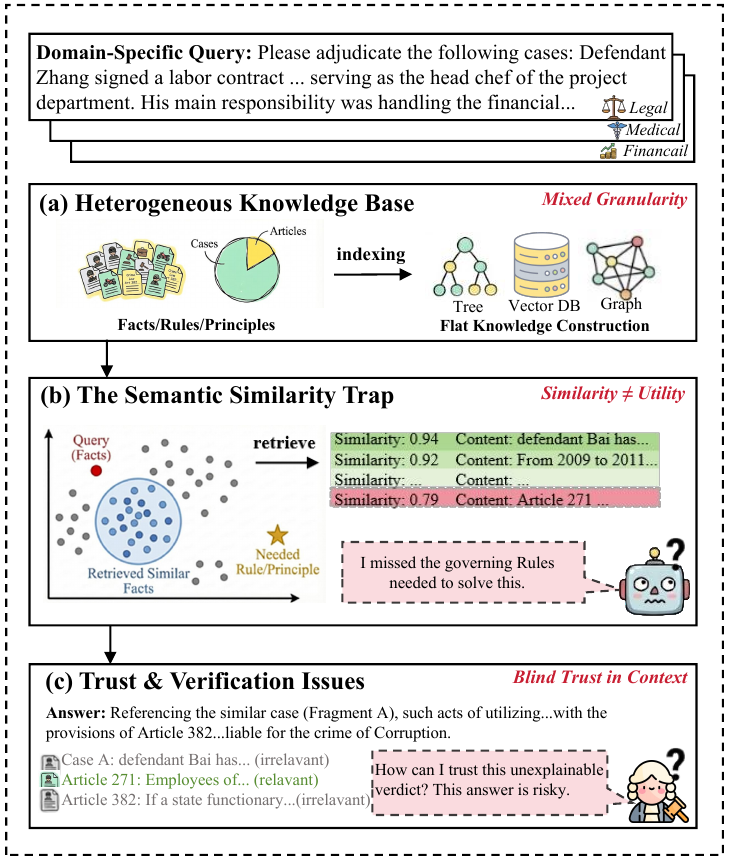}
    \caption{Challenges of Traditional RAG in Domain-Specific Tasks. (i) \textbf{Flat Graph Structure}: Struggles to handle heterogeneous documents. (ii) \textbf{Unverified Retrieval}: Contains excessive irrelevant information.}
    \label{fig:intro_comparision}
    \vspace{-5mm}
\end{figure}

\begin{figure}[t]
    \centering
    \includegraphics[width=\linewidth]{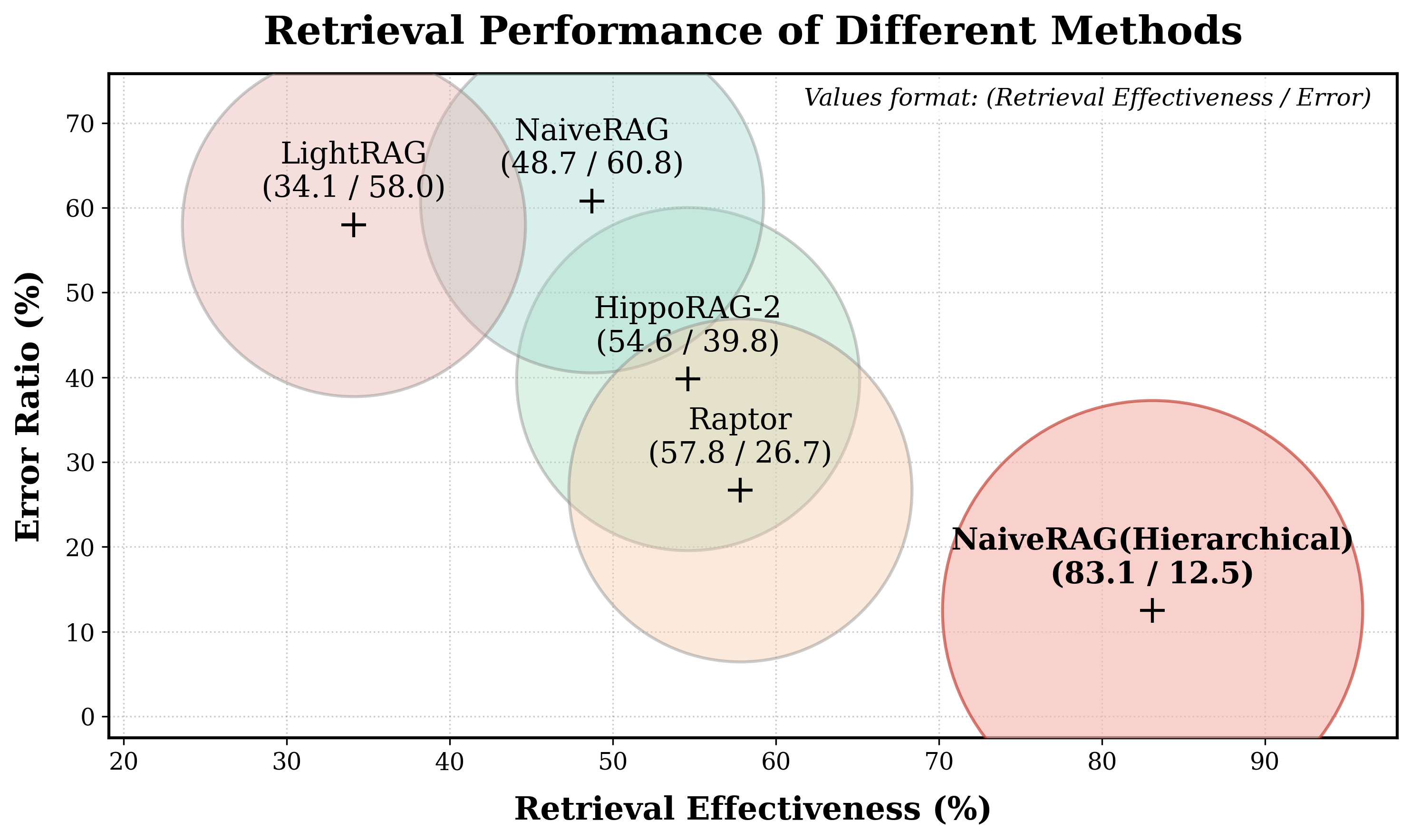}
    \caption{Retrieval performance comparison revealing that conventional RAG methods struggle with heterogeneous domain documents, suffering from high error rates and limited effectiveness. detailed experimental setup is introduced in Section~\ref{sec:prelimilary_hier} and Appendix~\ref{sec:Retrieval Effectiveness}.}
    \label{fig:pre_comparision}
    \vspace{-5mm}
\end{figure}

The rapid advancement of Large Language Models (LLMs), like GPT~\cite{achiam2023gpt}, Gemini~\cite{comanici2025gemini} and Qwen~\cite{yang2025qwen3} series, has driven significant progress in intelligent decision-making across various real-world tasks~\cite{zhao2023survey, naveed2025comprehensive}. However, deploying these models in specialized, knowledge-intensive fields like legal reasoning remains challenging due to the domain's demanding standards of rigor and reliability~\cite{lai2024large, hou2025large, siino2025exploring}. Domain-specific tasks necessitate a comprehensive understanding and multi-step reasoning across a vast knowledge base of specialized concepts, rigorous rules, and complex dependencies~\cite{wang2023survey, kim2025rethinking}, which requires strict logical reasoning and domain expertise that exceed the capabilities of general-purpose LLMs. While Supervised Fine-Tuning (SFT)~\cite{ouyang2022training, hu2022lora} on domain corpora enables models to internalize that expertise, this approach incurs substantial computational costs and often risks critical catastrophic forgetting in many real-world scenarios~\cite{yue2024lawllm, luo2025empirical}.

Recently, Retrieval-Augmented Generation (RAG)~\cite{lewis2020retrieval, borgeaud2022improving, li2025lexrag, zhang2025faithfulrag} offers a practical solution to adapt LLMs for specific domains. RAG systems enable LLMs to generate responses by leveraging not only their parametric knowledge but also real-time retrieved domain knowledge, thereby providing more accurate and reliable answers~\cite{mallen2023not, zhang2025faithfulrag}. However, standard RAG systems typically retrieve information based on semantic similarity~\cite{karpukhin2020dense, chen2024bge}, treating documents as independent text segments. This hinders complex multi-hop reasoning over hierarchical legal concepts and multiple documents, limiting effectiveness in legal analysis.



Graph-based Retrieval-Augmented Generation (GraphRAG)~\cite{edge2024local,zhang2025survey,xiang2025use,yang2026graph} advances this paradigm by organizing domain corpora into structured relational graphs. This structural awareness captures hierarchical relationships between different concepts, thereby enabling more precise retrieval and supporting the multi-hop reasoning required for complex queries. However, directly applying standard GraphRAG to the legal domain faces critical challenges (as illustrated in Figure \ref{fig:intro_comparision}): \ding{182} A flat graph structure cannot capture the multi-granular hierarchies present in legal corpora, which span factual details, applied rules, and abstract principles across legal cases, articles, and interpretations, thereby limiting accurate retrieval. \ding{183} Lack of verifiable, evidence-based reasoning. Traditional RAG passes retrieved context directly to an LLM without any verification. This ``retrieve-then-generate'' pipeline often results in opaque, error-prone reasoning.
In this paper, we propose LegalGraphRAG, a novel framework that synergizes graph-based retrieval with the multi-agent reasoning system for reliable legal reasoning. Specifically, LegalGraphRAG consists of two key components:  (i) Hierarchical legal graph (HierarGraph), which organizes legal knowledge into a hierarchical graph to effectively decouple historical cases, relevant statutes, and judicial interpretations, and (ii)  a multi-agent system for evidence-based reasoning~\cite{xi2023rise, xiangsystematic}, where the legal judgment process is structured as a transparent pipeline that retrieves, verifies, and reasons over graph-grounded evidence to produce interpretable decisions. Generally, our contributions are summarized as follows:
\begin{itemize}
\item We propose LegalGraphRAG, an evidence-based legal reasoning framework driven by a multi-agent system operating on a hierarchical knowledge graph, which address legal heterogeneity and ensure reliable reasoning.
\vspace{-3mm}
\item We design a hierarchical legal knowledge graph with Ontology, Fact, and Rule layers to model multi-granular legal knowledge and support accurate retrieval.
\vspace{-3mm}
\item We establish a multi-agent system for evidence-based reasoning that performs adjudication through a transparent pipeline of retrieval, validation, and synthesis, grounding judgments in verifiable evidence chains.
\vspace{-3mm}
\item Extensive experiments show that LegalGraphRAG consistently outperforms existing GraphRAG baselines and legal language models in accurate and trustworthy legal analysis.
\end{itemize}

\section{Problem Statement}
Complex legal reasoning is formulated as an open-ended generation task evaluating the decision-making capabilities of LLMs within the legal domain. Formally, given a criminal fact description $f$ and a defendant $d$, a LLM is tasked with predicting the applicable charges $y$. In this paper, we focus on integrating this reasoning framework with RAG to assess the model's ability to leverage external legal knowledge for judicial reasoning. This task can be organized into the following stages:

\noindent \textbf{Knowledge Organization.}
Given an offline corpus of legal documents $\mathcal{D}$, including historical cases, articles and interpretations we construct a domain-specific legal knowledge graph:

\begin{equation}\small
KG = \Phi(\mathcal{D}),
\end{equation}

where $\Phi(\cdot)$ denotes the organization function.

\noindent \textbf{Knowledge Retrieval.}
For a legal query characterized by criminal facts $f$ and a defendant $d$, we retrieve relevant evidence from $KG
$ to form a contextual reference:

\begin{equation}\small
\mathcal{C} = \mathcal{R}(f, d, KG),
\end{equation}

where $\mathcal{R}(\cdot)$ represents the retrieval operator.

\noindent \textbf{Judgment Generation.}
Finally, the legal judgment (e.g., charge) $y$ is inferred by reasoning over the query and retrieved evidence:

\begin{equation}\small
P(y \mid f, d, \mathcal{C}) = \mathcal{G}(f, d, \mathcal{C}),
\end{equation}

where $\mathcal{G}(\cdot)$ denotes the generator LLM.

\section{Preliminary Study}
Applying standard retrieval paradigms to the specialized, knowledge-intensive legal domain faces critical challenges due to the inherent structural complexity and rigorous standards of such fields. To illustrate these challenges, we conduct two preliminary experiments to empirically investigate the specific limitations of existing methods regarding knowledge granularity and generation quality.


\subsection{Investigation on Knowledge Granularity}\label{sec:prelimilary_hier}
Complex domain knowledge possesses an inherent hierarchy. In the legal context, this necessitates distinguishing between abstract statutory principles and concrete case facts. We hypothesize that standard retrieval strategies fail to distinguish between these semantic granularities because they treat all text segments in the same way. To verify this, we compare a \textit{Flat Strategy} against a naive \textit{Hierarchical Strategy} that explicitly segregates articles from case narratives (detailed in Appendix~\ref{sec:Retrieval Effectiveness}).

As illustrated in Figure~\ref{fig:pre_comparision}, the empirical results confirm our hypothesis. \textit{Flat Strategy} exhibit a distinct ``granularity bias'', frequently prioritizing high-frequency factual details due to surface-level semantic overlaps, often at the expense of essential abstract principles. Conversely, \textit{Hierarchical Strategy} aligns better with the domain's logical structure, improving retrieval performance by 25.3\%. This observation suggests that structural flatness constitutes a fundamental bottleneck for standard RAG when handling multi-granular knowledge.

\subsection{Investigation on Generation Quality}
Reliable domain reasoning demands not only information retrieval but also evidence verification. Real-world legal environments often contain documents that share similar keywords but differ fundamentally in their domain applicability. To simulate this realistic challenge, we conduct a test (detailed in Appendix~\ref{sec:context with irrelevant information}). Specifically, we inject legally plausible but factually irrelevant documents into the retrieval context to evaluate the model's ability to focus on relevant evidence.

\begin{table}[h]
\centering
\resizebox{\columnwidth}{!}{
\begin{tabular}{l|cc|cc|cc}
\toprule
& \multicolumn{2}{c|}{\cellcolor{headerblue}\textbf{Charge}} & \multicolumn{2}{c|}{\cellcolor{headerblue}\textbf{Articles}} & \multicolumn{2}{c}{\cellcolor{headerblue}\textbf{Term of Penalty}} \\
\cmidrule(lr){2-3} \cmidrule(lr){4-5} \cmidrule(lr){6-7}
\multirow{2}{*}{\textbf{Method}} & \textbf{ACC$\uparrow$} & \multirow{2}{*}{$\Delta$} & \textbf{ACC$\uparrow$} & \multirow{2}{*}{$\Delta$} & \textbf{MAE$\downarrow$} & \multirow{2}{*}{$\Delta$} \\
& (\%) & & (\%) & & (months) & \\
\midrule
RAG (Correct Context) & 42.8 & -- & 74.7 & -- & 24.3 & -- \\
RAG + 2 Irrelevant Docs & 34.9 & \cellcolor{red!10}$\downarrow$ 7.9 & 57.2 & \cellcolor{red!10}$\downarrow$ 17.5 & 27.7 & \cellcolor{blue!10}$\uparrow$ 3.4 \\
RAG + 4 Irrelevant Docs & 32.9 & \cellcolor{red!20}$\downarrow$ 9.9 & 51.1 & \cellcolor{red!20}$\downarrow$ 23.6 & 28.4 & \cellcolor{blue!20}$\uparrow$ 4.1 \\
RAG + 6 Irrelevant Docs & 29.8 & \cellcolor{red!30}$\downarrow$ 13.0 & 46.8 & \cellcolor{red!30}$\downarrow$ 27.9 & 31.7 & \cellcolor{blue!30}$\uparrow$ 7.4 \\
\bottomrule
\end{tabular}
}
\caption{Performance degradation under varying levels of simulated retrieval noise. ACC ($\uparrow$) denotes Accuracy for Charge and Articles prediction. MAE ($\downarrow$) represents Mean Absolute Error for Term of Penalty.}
\label{tab:noise_robustness}
\end{table}
\vspace{-2mm}

As summarized in Table~\ref{tab:noise_robustness}, standard RAG models exhibit significant sensitivity to context purity. The inclusion of irrelevant information precipitates a sharp performance drop. This observation shows that without a dedicated verification mechanism to filter irrelevant content, the model struggles to distinguish valid evidence from misleading information, which undermines reasoning reliability.

\subsection{Discussion and Motivation}
The findings from these two studies highlight fundamental limitations in applying standard RAG to complex domains: \ding{182} Flat retrieval mechanisms fail to navigate the hierarchical nature of domain knowledge (e.g., distinguishing rules from facts), resulting in biased context. \ding{183} The lack of an explicit verification step makes the system fragile to misleading information, which is unacceptable in rigorous fields like law. These insights motivate the design of LegalGraphRAG, which incorporates a \textit{Hierarchical Legal Graph} to resolve granularity conflicts and a \textit{Evidence-based Legal Reasoning} (Researcher-Auditor-Adjudicator) framework to enforce rigorous verification.

\begin{figure*}[t]
    \centering
    \includegraphics[width=\linewidth]{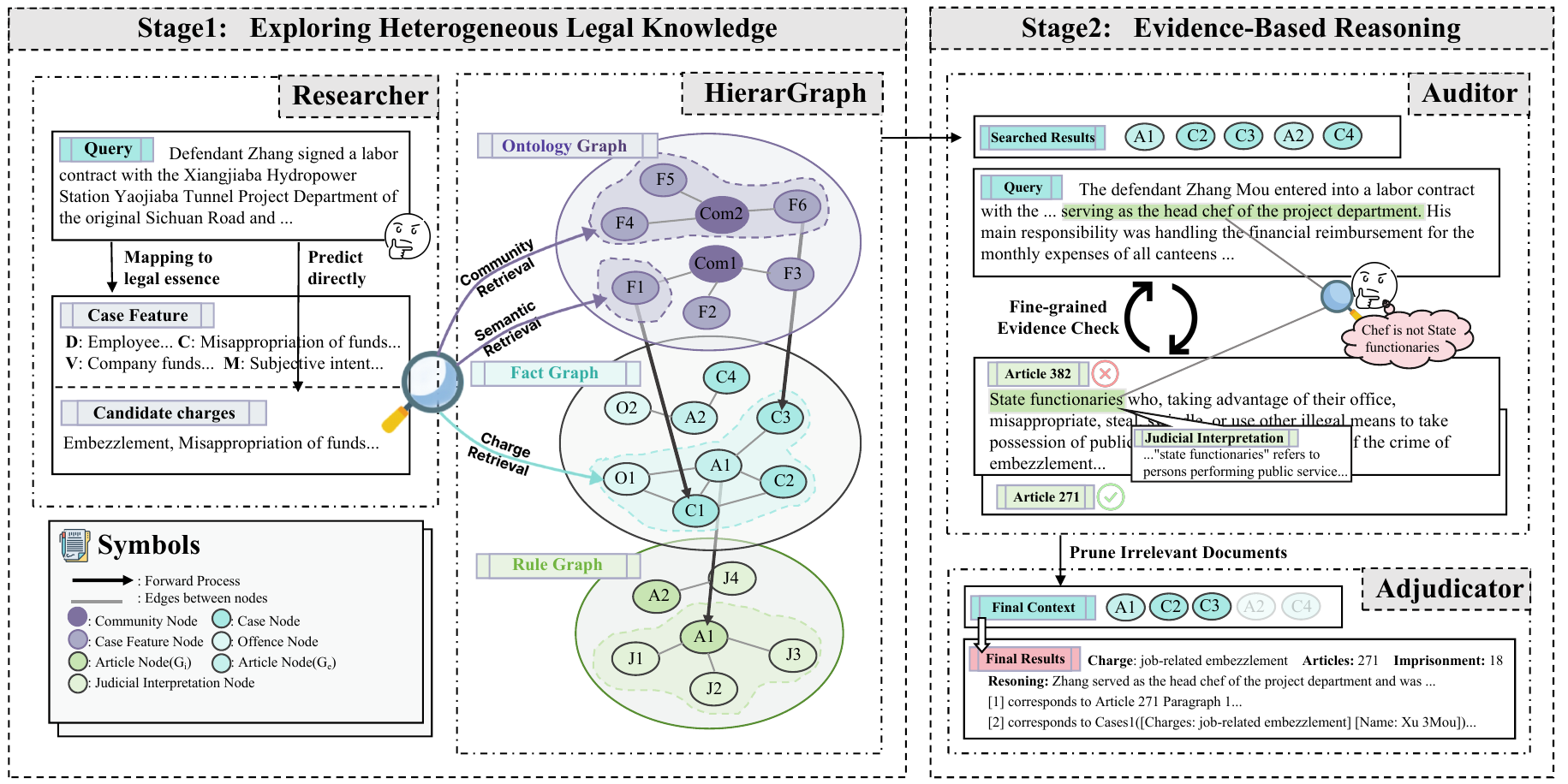}
    \caption{\textbf{The architecture of LegalGraphRAG.} The framework consists of two main phases: (1) \textbf{Hierarchical Knowledge Construction}, which builds a Hierarchical Legal Graph (HierarGraph) comprising an Fact Graph, Ontology Graph and Rule Graph to organize heterogeneous legal knowledge; and (2) \textbf{Evidence-based Legal Reasoning}, where a multi-agent system (Researcher, Auditor, and Adjudicator) performs structured retrieval, validation, and synthesis over the HierarGraph to generate interpretable legal decisions.}
    \label{fig:graph_structure}
    \vspace{-2mm}
\end{figure*}
\section{The Framework of LegalGraphRAG}
\subsection{Overview}

Traditional GraphRAG approaches face limitations in legal judgment due to the heterogeneous and multi-granular nature of legal corpora. To address this challenge, We propose LegalGraphRAG, an evidence-based legal reasoning framework driven by a multi-agent system operating on a hierarchical knowledge graph. The framework operates in two distinct phases: (i) \textbf{Hierarchical Knowledge Construction}, which organizes legal knowledge into a layered graph structure to effectively decouple historical cases, relevant statutes, and judicial interpretations, and (ii) \textbf{Evidence-based Legal Reasoning}, structures the legal judgment process as a transparent pipeline that retrieves, verifies, and reasons over graph-grounded evidence to produce interpretable decisions. The whole framework is illustrated in Figure~\ref{fig:graph_structure}.

\subsection{Hierarchical Knowledge Construction}
Legal reasoning involves heterogeneous information sources, including historical cases, abstract legal articles, and interpretations. Employing a flat storage structure is not enough to handle the inherent structural differences of these data sources, leading to disorganized information and inefficient retrieval. To address this challenge, we construct a Hierarchical Legal Graph (HierarGraph) $\mathcal{H}$ that organizes legal knowledge into distinct semantic layers, enabling explicit differentiation among legal concepts and providing a structured basis for reliable reasoning. The HierarGraph is composed of three specialized subgraphs:

\noindent \textbf{Fact Graph ($\mathcal{G}_{fac}$)}, which serves as a structured collection of verified legal precedents, providing the essential factual basis for ensuring legally grounded judgments. Accordingly, $\mathcal{G}_{\text{fac}}$ models the natural structure of legal documents by explicitly connecting \textit{Cases} ($\mathcal{C}$), \textit{Articles} ($\mathcal{A}$), and \textit{Offense} ($\mathcal{O}$) nodes. Relationships are established via $e_{ca}$, linking a case $c$ to its cited article $a$, and $e_{co}$, linking a case $c$ to its convicted offense $o$. This structure provides the factual granularity required for evidence gathering. Formally, it is defined as:
\begin{equation}\small
\mathcal{G}_{fac}
= (\mathcal{V}_{fac}, \mathcal{E}_{fac})
= \left(
    \bigl\{ c_i, a_i, o_i \bigr\}_{i=1}^{|\mathcal{G}_\mathsf{fac}|}\right).
\end{equation}

\noindent \textbf{Ontology Graph ($\mathcal{G}_{ont}$)}, which bridges the semantic gap and mitigates noise by abstracting case features. $\mathcal{G}_{ont}$ distills raw narratives containing instance-specific details (e.g., dates and locations) into a purified semantic space that reflects the ``legal essence''. Specifically, we design a domain-specific legal ontology based on legal theory~\cite{rüthers2013rechtstheorie}, encompassing four key dimensions: \textit{Defendant Attributes}, \textit{Criminal Behaviors}, \textit{Victim Characteristics} and \textit{Subjective Mental States}. Keywords and entities are extracted and aligned with these properties to form structured embeddings, serving as indices for \textit{Case Feature Nodes} ($\mathcal{F}$).~\label{sec:method_ontology}

To reveal hidden connections between different cases, we employ the k-Nearest Neighbors (k-NN) algorithm to connect nodes with high semantic similarity. We then  apply the Leiden algorithm~\cite{traag2019louvain} to group related cases into communities, each treated as a \textit{Community Node} ($\mathcal{K}$). Each $k$ contains the summarized information of the cases inside it, facilitating hierarchical retrieval that navigates from broad contexts to specific details. Formally, this subgraph is defined as:
\begin{equation}\small
\mathcal{G}_{ont}
= (\mathcal{V}_{ont}, \mathcal{E}_{ont})
= \left(
    \bigl\{ c_i, k_j \bigr\}_{i=1,j=1}^{|\mathcal{G}_{ont}|}
\right) .
\end{equation}

\noindent \textbf{Rule Graph ($\mathcal{G}_{rul}$)}, which resolves statutory ambiguities by systematically linking \textit{Articles} ($\mathcal{A}$) with its corresponding \textit{Judicial Interpretations} ($\mathcal{J}$). This explicit alignment establishes the contextual grounding necessary for precise legal reasoning.

Moreover, applying the correct article often depends on specific conditions. A small difference can lead to a completely different judgment for the same crime. (e.g. whether the defendant is an adult or a minor). Simple semantic matching often fails to distinguish these subtle differences. To address this, we equip each $a$ with a \textit{Diagnostic Checklist} ($\mathcal{D}$). This mechanism breaks down complex legal rules into specific verification steps. Formally, this subgraph is defined as: 
\vspace{-2mm}

\begin{equation}\small
\mathcal{G}_{rul}
= (\mathcal{V}_{rul}, \mathcal{E}_{rul})
= \left(
\bigl\{ a_i, j_i \bigr\}_{i=1}^{|\mathcal{G}_{rul}|} \right) . 
\end{equation}
where
\begin{equation}\small
\mathcal{D}(a_i) = \{d_1,\dots,d_{|C|}\}
\end{equation}

By integrating these three layers, HierarGraph $\mathcal{H}$ transforms heterogeneous legal corpora into a structured ecosystem. This architecture directly addresses the limitations of flat retrieval by offering multi-granular support for following evidence-based legal reasoning. The detailed construction procedures are provided in Appendix~\ref{sec:construct detail}.

\subsection{Evidence-based Legal Reasoning}
To leverage the multi-granular knowledge encoded in our HierarGraph, we propose a multi-agent system for evidence-based reasoning, in which specialized agents sequentially traverse the graph to perform evidence retrieval, validation, and synthesis. Specifically, the workflow consists of three agents:1)  \textit{Researcher}, 2) \textit{Auditor}, and 3) \textit{Adjudicator}. Through structured graph traversal and logical analysis, the framework resolves the raw case query by constructing a final, verifiable judgment.

\subsubsection{Evidence Retrieval}
A reliable evidence-based reasoning process begins with grounding a raw case description in relevant legal evidence. To this end, \textit{Researcher} perform structured evidence retrieval over the $\mathcal{G}_{ont}$ and the $\mathcal{G}_{\text{fac}}$, transforming unstructured case narratives into a coherent set of related  \textit{Cases} ($\mathcal{C}$) and \textit{Articles} ($\mathcal{A}$).

Specifically, the Researcher aligns the case description with the ontological dimensions defined in Section~\ref{sec:method_ontology}. Based on these features, We formulate the evidence retrieval process $\mathcal{R}(q)$ as the union of three operators, where $q$ is the legal query:
\begin{equation}\small
\mathcal{R}(q) = \mathcal{R}_{\text{sem}}(q) \cup \mathcal{R}_{\text{com}}(q) \cup \mathcal{R}_{\text{chg}}(q)
\end{equation}
First, we employ \textit{Semantic Match Retrieval} to locate direct evidence via semantic similarity, where $\phi(\cdot)$ denotes ontology-aligned embeddings:
\begin{equation}\small 
\mathcal{R}_{\text{sem}}(q) = \operatorname*{Top-k}_{c \in \mathcal{G}_{ont}} \text{sim}\bigl(\phi(q), \phi(c)\bigr) 
\end{equation}

Next, to capture structural context, we conduct \textit{Community Expansion Retrieval}. We first identify the top-ranked communities by topic $\mathcal{S}_{\mathcal{K}}$ aligned with the query, and then retrieve the most similar cases within these communities:
\begin{equation}\small
\begin{split}
    \mathcal{K}^* &= \operatorname*{argmax}_{\mathcal{K} \in \mathcal{G}_{ont}} \text{sim}\bigl(\phi(q), \phi(\mathcal{K})\bigr) \\
    \mathcal{R}_{\text{com}}(q) &= \operatorname*{Top-k}_{c \in \mathcal{K}^*} \text{sim}\bigl(\phi(q), \phi(c)\bigr)
\end{split}
\end{equation}

Finally, we implement \textit{Charge-Anchored Retrieval} to anchor the legal basis by collecting cases linked to inferred charges. Here, $\mathcal{O}(q)$ denotes the set of predicted charges and $\mathcal{N}$ represents the neighboring cases connected to charge $o$ in $\mathcal{G}_{fac}$:
\begin{equation}\small 
\mathcal{R}_{\text{chg}}(q) = \bigcup_{o \in \mathcal{O}(q)} \mathcal{N}_{\mathcal{G}_{fac}}(o) \end{equation}

The specific retrieval algorithms and parameter settings are detailed in Appendix \ref{sec:researcher detail}.

\begin{table*}[t!]
\centering
\resizebox{\textwidth}{!}{
\begin{tabular}{ll|cccccccc|cccccccc|cc}
\toprule
& & \multicolumn{8}{c|}{\cellcolor{headerblue}\textbf{CAIL}} & \multicolumn{8}{c|}{\cellcolor{headerpurple}\textbf{CMDL}} & \multicolumn{2}{c}{\textbf{Average}} \\
\cmidrule(lr){3-10} \cmidrule(lr){11-18}
\multirow{3}{*}{\textbf{Model}} & \multirow{3}{*}{\textbf{Size}} & \multicolumn{2}{c}{\textbf{Public Safety}} & \multicolumn{2}{c}{\textbf{Economic}} & \multicolumn{2}{c}{\textbf{Social Order}} & \multicolumn{2}{c|}{\textbf{Person Rights}} & \multicolumn{2}{c}{\textbf{Public Safety}} & \multicolumn{2}{c}{\textbf{Economic}} & \multicolumn{2}{c}{\textbf{Social Order}} & \multicolumn{2}{c|}{\textbf{Person Rights}} & \multirow{3}{*}{\textbf{All}} & \multirow{3}{*}{$\Delta$} \\
& & \multicolumn{2}{c}{\textbf{ACC / F1}} & \multicolumn{2}{c}{\textbf{ACC / F1}} & \multicolumn{2}{c}{\textbf{ACC / F1}} & \multicolumn{2}{c|}{\textbf{ACC / F1}} & \multicolumn{2}{c}{\textbf{ACC / F1}} & \multicolumn{2}{c}{\textbf{ACC / F1}} & \multicolumn{2}{c}{\textbf{ACC / F1}} & \multicolumn{2}{c|}{\textbf{ACC / F1}} & & \\
\cmidrule(lr){3-4} \cmidrule(lr){5-6} \cmidrule(lr){7-8} \cmidrule(lr){9-10} \cmidrule(lr){11-12} \cmidrule(lr){13-14} \cmidrule(lr){15-16} \cmidrule(lr){17-18}
\midrule
\multicolumn{20}{c}{\textbf{Open-Source Models}} \\
\midrule
Qwen-2.5-7B-Instruct & 7B-Inst & 24.0 & 45.8 & 23.1 & 42.5 & 22.9 & 36.7 & 27.4 & 46.0 & 25.8 & 32.4 & 28.7 & 35.8 & 27.2 & 42.1 & 32.8 & 49.6 & 26.7 & \cellcolor{darkgreen}$\uparrow$ 22.8 \\
Qwen-3-8B & 8B-Inst & 31.7 & 49.2 & 25.8 & 42.7 & 26.3 & 39.8 & 27.6 & 47.8 & 44.0 & 52.3 & 44.7 & 53.1 & 42.7 & 51.9 & 53.0 & 57.7 & 35.2 & \cellcolor{midgreen}$\uparrow$ 19.9 \\
Internlm3-8b-instruct & 8B-Inst & 29.8 & 49.1 & 26.7 & 42.0 & 25.2 & 34.3 & 28.1 & 47.3 & 25.4 & 32.1 & 35.7 & 37.0 & 27.5 & 36.2 & 34.1 & 53.6 & 26.6 & \cellcolor{darkgreen}$\uparrow$ 22.9 \\
Glm-4-9b-chat & 9B-Inst & 18.4 & 33.7 & 19.7 & 36.1 & 15.8 & 32.1 & 26.0 & 44.5 & 23.5 & 34.2 & 23.6 & 40.8 & 19.1 & 37.0 & 41.5 & 47.0 & 21.2 & \cellcolor{darkgreen}$\uparrow$ 28.2 \\
\midrule
\multicolumn{20}{c}{\textbf{Advanced Models}} \\
\midrule
GPT-4o-mini & $\sim$8B & 19.7 & 35.5 & 19.6 & 33.3 & 15.5 & 35.2 & 29.0 & 46.3 & 18.0 & 28.0 & 22.7 & 31.9 & 21.9 & 32.0 & 35.9 & 50.3 & 28.4 & \cellcolor{darkgreen}$\uparrow$ 21.1 \\
DeepSeek-V3.1 & $\sim$200B & 31.0 & \underline{51.3} & 29.0 & 48.4 & 29.8 & \underline{50.2} & \underline{35.2} & 54.8 & 35.0 & \underline{64.0} & 54.7 & \underline{62.7} & \underline{58.2} & 61.9 & 62.5 & \underline{71.6} & 42.8 & \cellcolor{lightgreen}$\uparrow$ 6.7 \\
\midrule
\multicolumn{20}{c}{\textbf{Legal Specific Methods}} \\
\midrule
DISC-LawLLM-7B & 7B-Inst & 40.1 & 50.9 & 31.0 & \underline{51.5} & \underline{34.8} & 47.7 & 34.5 & \underline{56.0} & 49.7 & 53.6 & 39.6 & 52.1 & 30.3 & 49.5 & 48.4 & 63.3 & 30.3 & \cellcolor{midgreen}$\uparrow$ 19.1 \\
ADAPT & 7B-Inst & 38.7 & 43.7 & \underline{32.7} & 43.4 & 27.6 & 41.7 & 35.2 & 50.7 & 54.5 & 58.8 & \underline{57.1} & 59.4 & 40.9 & 43.4 & 61.5 & 62.1 & 42.8 & \cellcolor{lightgreen}$\uparrow$ 6.7 \\
Legal $\Delta$ & 7B-Inst & \underline{40.8} & 50.6 & 25.1 & 37.4 & 32.1 & 43.7 & 34.1 & 53.6 & \underline{58.3} & 61.5 & 51.8 & 55.8 & 50.2 & 54.8 & \underline{65.8} & 64.4 & 42.4 & \cellcolor{lightgreen}$\uparrow$ 7.1 \\
\midrule
\multicolumn{20}{c}{\textbf{RAG Based Methods}} \\
\midrule
Naive RAG & 8B-Inst & 31.0 & 45.7 & 24.4 & 38.7 & 28.1 & 38.4 & 34.5 & 46.8 & 45.8 & 57.3 & 44.8 & 55.2 & 46.8 & 58.5 & 49.6 & 57.8 & 33.3 & \cellcolor{midgreen}$\uparrow$ 16.1 \\
G-retriever & 8B-Inst & 33.8 & 48.0 & 26.0 & 39.8 & 23.8 & 39.3 & 32.6 & 50.1 & 36.8 & 40.0 & 42.5 & 48.8 & 45.3 & 50.7 & 46.2 & 52.4 & 34.4 & \cellcolor{midgreen}$\uparrow$ 13.2 \\
LightRAG & 8B-Inst & 20.4 & 43.6 & 21.7 & 42.5 & 19.0 & 42.5 & 26.9 & 50.6 & 37.9 & 50.1 & 43.2 & 45.1 & 44.2 & 51.3 & 43.7 & 46.9 & 30.5 & \cellcolor{midgreen}$\uparrow$ 19.0 \\
RAPTOR & 8B-Inst & 34.6 & 50.4 & 31.6 & 43.9 & 32.1 & 45.6 & 32.4 & 45.7 & 53.8 & 62.6 & 53.6 & 60.1 & 52.5 & \underline{62.8} & 52.1 & 66.9 & \underline{43.1} & \cellcolor{lightgreen}$\uparrow$ 6.3 \\
HippoRAG2 & 8B-Inst & 34.5 & 38.2 & 24.0 & 33.5 & 28.8 & 35.0 & 31.0 & 36.3 & 53.5 & 56.5 & 50.6 & 52.7 & 53.5 & 55.0 & 62.4 & 62.8 & \underline{43.1} & \cellcolor{lightgreen}$\uparrow$ 6.3 \\
\textbf{LegalGraphRAG}~(Ours) & 8B-Inst & \textbf{42.9} & \textbf{54.3} & \textbf{38.5} & \textbf{53.6} & \textbf{37.6} & \textbf{51.1} & \textbf{37.2} & \textbf{58.3} & \textbf{65.5} & \textbf{66.5} & \textbf{59.8} & \textbf{65.1} & \textbf{58.5} & \textbf{63.7} & \textbf{70.1} & \textbf{72.7} & \textbf{49.5} & -- \\
\bottomrule
\end{tabular}
    }
\caption{\textbf{Performance comparison on CAIL and CMDL.} We employ Qwen3-8B as the default backbone model. The best results are highlighted in \textbf{bold}, and the second-best are \underline{underlined}. We visualize the gains of LegalGraphRAG over each baseline in the \colorbox{lightgreen}{$\Delta$ columns}.}
\label{tab:main_results}
\end{table*}

\subsubsection{Evidence Validation}
Given the candidate evidence retrieved in the Evidence Retrieval, this stage focuses on validating whether the case facts genuinely satisfy the conditions required by the law, rather than relying on surface-level semantic relevance.

Specifically, for each candidate article, we verify its applicability by evaluating the case facts using the associated \textit{Diagnostic Checklist} and \textit{Judicial Interpretations} encoded in the $\mathcal{G}_{\text{rul}}$. The verification outcomes are then aggregated to produce a definitive applicability judgment for each article.

Based on these judgments, \textit{Auditor} filters the retrieval subgraph by pruning inapplicable articles and their associated case and charge nodes. Finally, it organizes the remaining nodes into a legally consistent and evidence-supported subgraph, which serves as a validated knowledge basis for subsequent decision-making. Further implementation details can be found in Appendix \ref{sec:auditor detail}.




\subsubsection{Evidence Synthesis}
In the final stage, the validated evidence produced in the previous steps is synthesized to derive a legally grounded judgment. Based on the verified subgraph, \textit{Adjudicator} integrates the confirmed articles ($\mathcal{A}^f$), cases ($\mathcal{C}^f$), and offense information ($\mathcal{O}^f$) to determine the applicable charges and their statutory basis. This process is formulated as:

\begin{equation}\small
    \mathcal{J} = Adjudicator (q \oplus \mathcal{A}^f \oplus \mathcal{C}^f \oplus \mathcal{O}^f)
\end{equation}
Crucially, the judgment is not produced as a direct verdict. Instead, it is accompanied by explicit citations to the statutory articles and judicial interpretations used in the reasoning process, ensuring that every conclusion is directly traceable to verified evidence in the HierarGraph.



Overall, LegalGraphRAG formulates legal judgment as a transparent, evidence-based reasoning pipeline rather than a black-box generation process. Through sequential evidence grounding, validation, and synthesis, the system enforces stepwise verification and ensures that every conclusion is explicitly derived from and supported by verified legal evidence, resulting in reliable judicial decisions.

\begin{figure*}[t]
    \centering
    \includegraphics[width=0.99\linewidth]{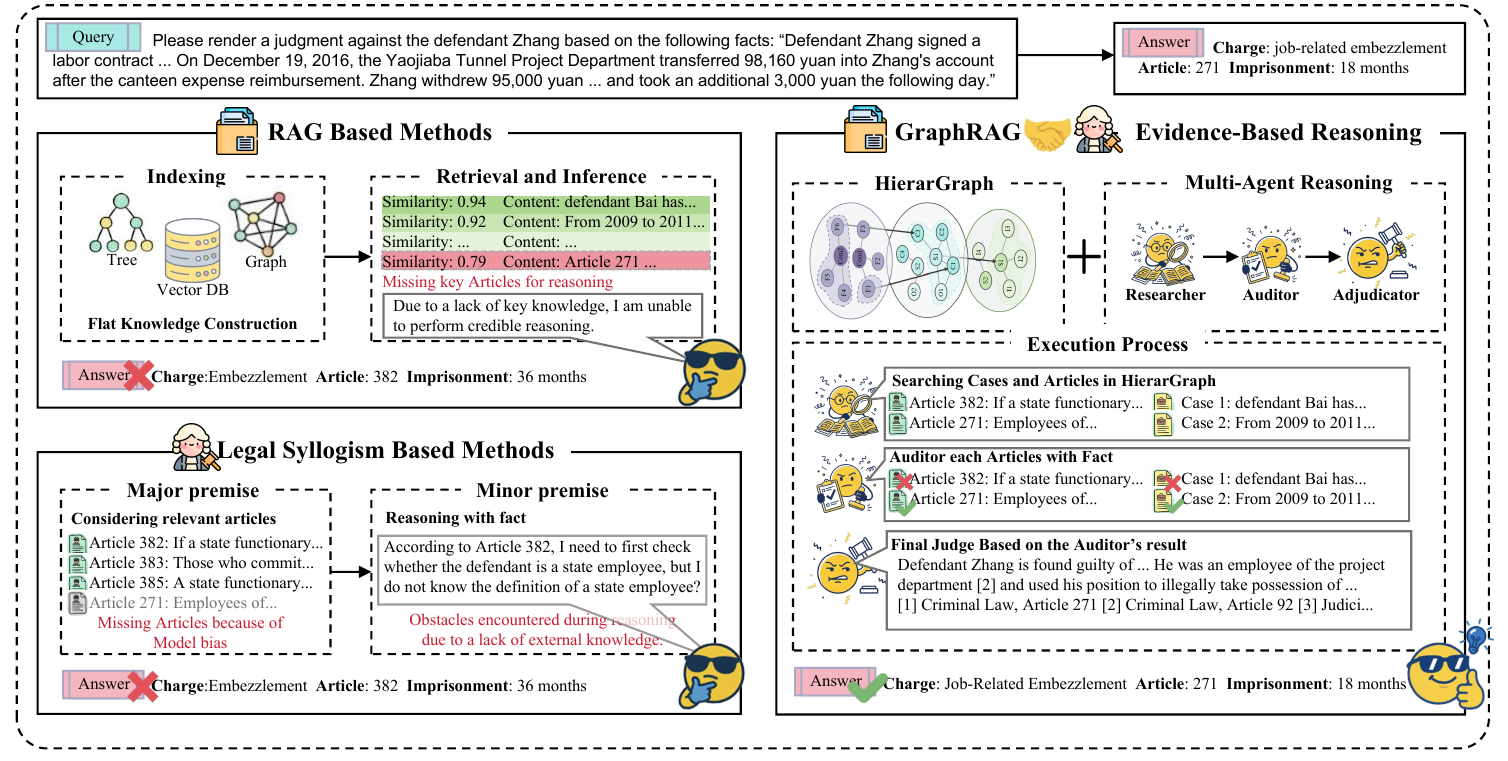}
    \caption{\textbf{A comparative case study} illustrating the reasoning trajectories of different methods. While Naive RAG fails due to missing legal articles and syllogism-based methods struggle with ambiguities, LegalGraphRAG derives the correct judgment. By leveraging the HierarGraph and Evidence-based Legal Reasoning, our framework demonstrates transparency and reliability, providing a verifiable reasoning chain grounded in legal evidence.}
    \label{fig:case_study}
    \vspace{-2mm}
\end{figure*}


\section{Experiment}
This section presents a comprehensive evaluation of LegalGraphRAG on two legal judgment benchmarks. Our experiments are designed to answer the following three questions. \textbf{Q1 (Generation Accuracy)}: Does LegalGraphRAG outperform SOTA GraphRAG methods and leading legal-domain LLMs in generation quality? \textbf{Q2 (Case Study)}: How does LegalGraphRAG handle specific legal cases, and does it provide more interpretable outputs compared to baselines? \textbf{Q3 (Ablation Study)}: What is the contribution of each core component to the final performance of LegalGraphRAG? More additional experiments are provided in Appendix~\ref{sec:additional_exp}
\subsection{Experiment Setup}

\noindent \textbf{Datasets}\quad We evaluate on two widely used legal benchmarks: CAIL2018~\cite{xiao2018cail2018} and CMDL~\cite{huang2024cmdl}, covering diverse criminal sub-fields such as Public Safety, Social Order, Economic Offenses, and Person Rights. The retrieval knowledge base is built from a collection of authoritative legal sources, including case datasets and statutory texts. Further dataset details are provided in Appendix \ref{sec:Benchmark Dataset} \& \ref{sec:Corpus}.

\noindent \textbf{Baselines}\quad To ensure a comprehensive evaluation, we categorize our comparative experiments into four distinct groups: (i) Open-Source Models, utilizing Qwen-series~\cite{yang2025qwen3}, InternLM~\cite{fei2025internlm} and GLM~\cite{glm2024chatglm} as foundational backbones; (ii) Advanced Models, represented by GPT-4o-mini~\cite{achiam2023gpt} and DeepSeek-V3.1~\cite{liu2024deepseek}. (iii) Legal-Specific Methods, which include domain-specialized approaches such as Disc-LLM~\cite{yue2024lawllm}, Legal$\Delta$~\cite{dai2025legal}, and ADAPT~\cite{deng2024enabling}; and (iv) RAG-Based Methods, encompassing Naive RAG and advanced graph-augmented strategies like G-retriever~\cite{he2024g}, RAPTOR~\cite{sarthi2024raptor}, LightRAG~\cite{guo2024lightrag}, and HippoRAG2~\cite{gutierrez2025rag}. Detailed configurations are provided in Appendix \ref{sec:Baseline Details}.

\begin{figure}[t]
    \centering
    \includegraphics[width=0.95\linewidth]{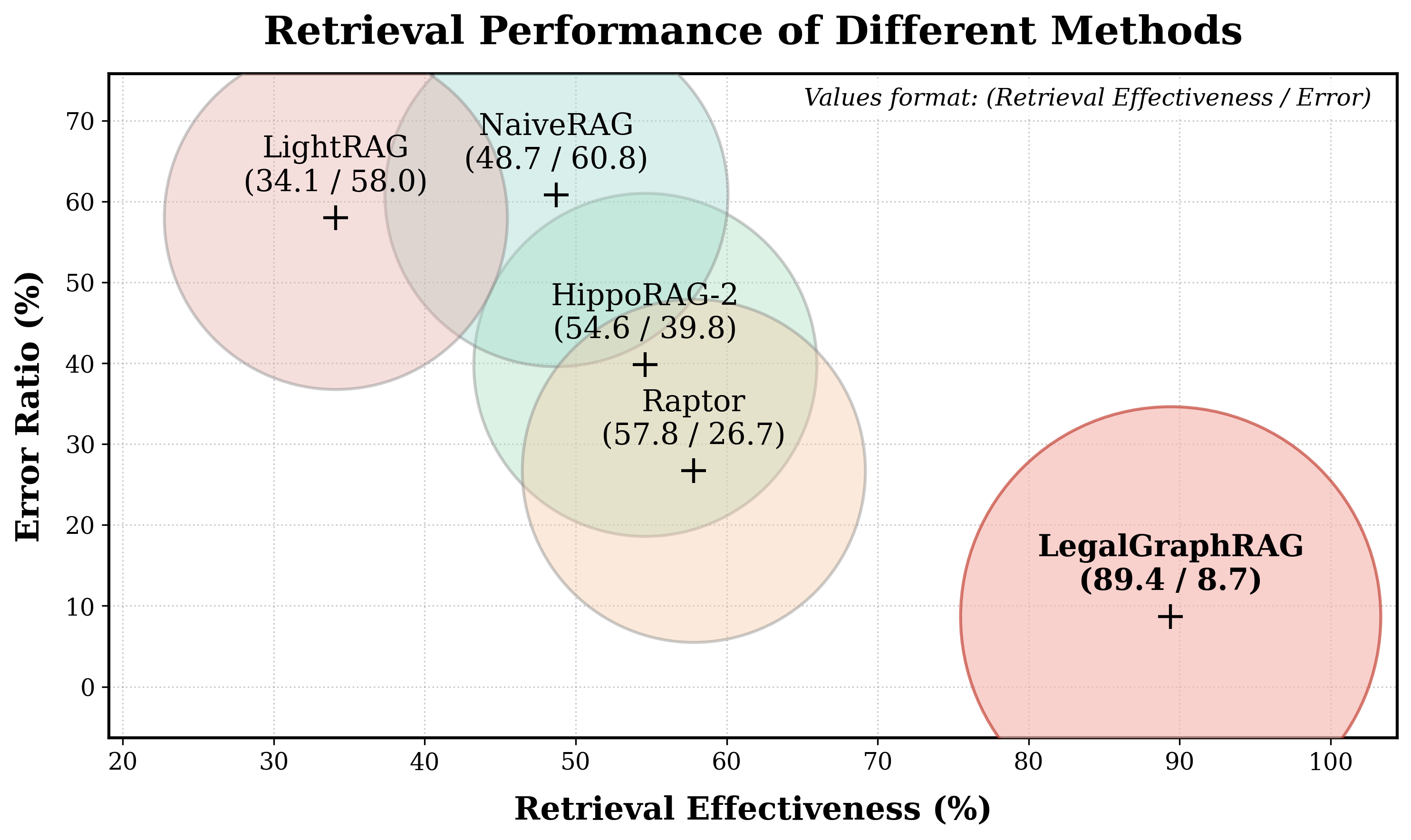}
    \vspace{-3mm}
    \caption{\textbf{Retrieval Performance Comparison.} LegalGraphRAG demonstrates superior retrieval effectiveness and significantly lower error ratios compared to conventional flat graph baselines.}
    \label{fig:retrieval_comparison}
    \vspace{-5mm}
\end{figure}

\noindent \textbf{Evaluation Metrics}\quad We employ Accuracy and Micro-F1 score to evaluate prediction performance. Detailed definitions are provided in Appendix~\ref{sec:Evaluation Metrics}.

\noindent \textbf{Implementation Details}\quad We utilize GPT-4o-mini for graph construction and BGE-m3~\cite{chen2024bge} for embedding generation. Various LLMs serve as backbone models for the reasoning phase. We employ Qwen3-8B~\cite{yang2025qwen3} as the default backbone model for our main experiments. Full hyperparameter settings and hardware specifications are detailed in Appendix \ref{sec:LegalGraphRAG Setup}.

\subsection{Generation Accuracy (Q1)}
To address Q1, we evaluate LegalGraphRAG against SOTA RAG methods and specialized legal LLMs on two legal judgment datasets. The primary comparison results for charge prediction are reported in Table \ref{tab:main_results}, with extended analyses in Tables \ref{tab:more_results_closed_models_CMDL}, \ref{tab:more_results_laws}, and \ref{tab:more_results_terms} in Appendix. We summarize the key observations below.

\noindent\textbf{Obs.1. LegalGraphRAG consistently outperforms baselines in legal datasets.} Our method achieves the best results on most evaluation metrics across both datasets. Notably, LegalGraphRAG delivers significant improvements ranging from 6.3\% to 19.1\% over the strongest baselines. Unlike standard GraphRAG methods that struggle in the legal domain, our approach effectively structures heterogeneous knowledge, thereby enhancing legal reasoning capabilities and improving charge prediction accuracy overall.

\begin{figure}[t]
    \centering
    \includegraphics[width=\linewidth]{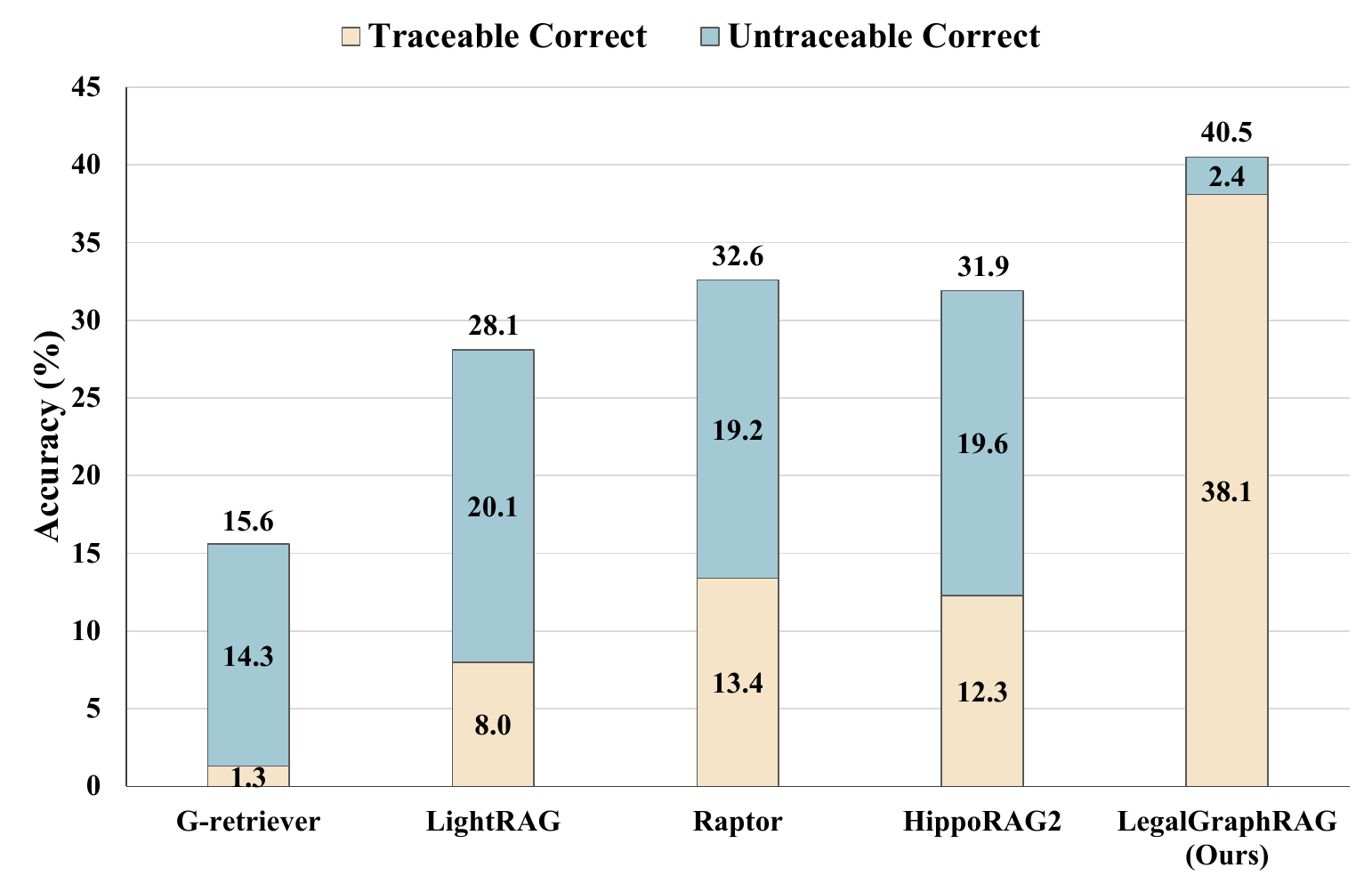}
    \caption{\textbf{Reliability Analysis.} LegalGraphRAG significantly increases the proportion of \textcolor[HTML]{F5E4C8}{\textbf{Traceable Correct}} samples, effectively minimizing \textcolor[HTML]{A3C9D5}{\textbf{Untraceable Correct}} predictions where the answer is correct but lacks supporting evidence in the retrieved context.}
    \vspace{-5mm}
    \label{fig:reliability_analysis}
    \vspace{-1mm}
\end{figure}

\noindent\textbf{Obs.2. LegalGraphRAG substantially surpasses existing specialized legal LLMs.} Our approach outperforms Legal~$\Delta$ and ADAPT by an average of 7.1\% and 6.7\%, respectively. Moreover, as shown in Table~\ref{tab:more_results_closed_models_CMDL} in Appendix, LegalGraphRAG integrates flexibly with different backbone models, achieving a peak performance of 78.7\% on CMDL when combined with strong backbones. This demonstrates strong adaptability and robust reasoning compared to specialized legal-domain baselines.

\subsection{Case Study (Q2)}
To demonstrate the superior interpretability of our framework, we present a qualitative analysis of a representative criminal case in Figure \ref{fig:case_study}. More cases are provided in Appendix \ref{sec:Extended Case Examples}.

\noindent\textbf{Obs.3. LegalGraphRAG retrieves significantly more relevant and comprehensive evidence.} As illustrated in Figure \ref{fig:retrieval_comparison}, conventional flat graph structures (e.g., HippoRAG2) struggle to handle heterogeneous legal documents, often failing to capture essential statutes. This structural limitation leads to fragmented context. In contrast, our hierarchical organization effectively structures legal knowledge, ensuring that the retrieved context is sufficient to support robust reasoning.

\noindent\textbf{Obs.4. LegalGraphRAG guarantees decision traceability through rigorous evidence grounding.} 
While baseline models often achieve correct predictions, our reliability analysis (Figure \ref{fig:reliability_analysis}) reveals a critical issue of ``unsupported correctness'', where the model predicts the right charge but fails to retrieve the necessary supporting evidence. This implies that the prediction is not supported by relevant evidence or a valid reasoning chain. LegalGraphRAG significantly increases the ratio of ``Traceable Correct'' samples (defined in Appendix \ref{sec:reliability_detail}). By enforcing strict verification, our system ensures that every statute cited in the judgment is explicitly present in the retrieved context, transforming opaque predictions into transparent, traceable decisions.

\begin{table}[t]
\centering
\resizebox{0.8\columnwidth}{!}{ 
\begin{tabular}{l|cc}
\toprule
& \multicolumn{2}{c}{\cellcolor{headerblue}\textbf{CAIL}} \\
\cmidrule{2-3}
\textbf{Settings} & \textbf{ACC} & \textbf{$\Delta$} \\
\midrule
\textbf{LegalGraphRAG} (Full) & \textbf{40.9} & -- \\
\midrule
w/o HierarGraph & 33.7 & \cellcolor{red!30}$\downarrow$ 7.2 \\
\midrule
w/o Researcher & 36.9 & \cellcolor{red!20}$\downarrow$ 4.0 \\
\quad w/o Semantic Match & 39.1 & \cellcolor{red!10}$\downarrow$ 1.8 \\
\quad w/o Community Exp. & 38.5 & \cellcolor{red!10}$\downarrow$ 2.4 \\
\quad w/o Charge-Anchored & 39.3 & \cellcolor{red!10}$\downarrow$ 1.6 \\
\midrule
w/o Auditor & 37.5 & \cellcolor{red!20}$\downarrow$ 3.4 \\
\bottomrule
\end{tabular}
}
\caption{\textbf{Ablation study} of LegalGraphRAG components on the CAIL dataset. Results underscore the indispensable role of the HierarGraph for knowledge organization and the synergy between the Researcher and Auditor agents in ensuring reasoning accuracy.}
\label{tab:ablation_study}
\vspace{-5mm}

\end{table}

\subsection{Ablation Study (Q3)}
To quantify the impact of each component, we performed a systematic ablation study by removing specific modules from the full LegalGraphRAG framework. Results are detailed in Table \ref{tab:ablation_study}.

\noindent\textbf{Obs.5. Hierarchical structure is the cornerstone of performance.}
Removing the hierarchical graph (\textit{w/o HierarGraph}) causes the sharpest accuracy drop of 7.2\%. This confirms that separating concrete facts from abstract rules into distinct granular levels is essential, providing structural precision that flat indexing lacks.

\noindent\textbf{Obs.6. The multi-agent workflow guarantees reasoning reliability.}
Excluding the \textit{Researcher} and \textit{Auditor} degrades accuracy by 4.0\% and 3.4\%, respectively. This validates their synergistic roles: the \textit{Researcher} maximizes evidence coverage through diverse retrieval strategies, while the \textit{Auditor} enforces rigorous verification, ensuring only validated evidence supports the judgment.


\section{Conclusion}
\vspace{-2mm}
In conclusion, we have presented LegalGraphRAG, an evidence-based legal reasoning framework that addresses the critical challenges of legal heterogeneity and reasoning reliability. By integrating a hierarchical knowledge graph with a collaborative multi-agent system, our approach transforms the legal reasoning process into a transparent pipeline of retrieval, verification, and synthesis. Extensive experiments on legal judgment benchmarks validate that LegalGraphRAG establishes a new state-of-the-art, significantly advancing accurate and trustworthy AI for reliable and complex legal analysis.

\section*{Limitation}
While LegalGraphRAG demonstrates significant proficiency in processing textual legal documents and statutes, its current scope is confined to unimodal textual inputs. Real-world judicial proceedings, however, often rely on a heterogeneity of evidence types, including crime scene photography, surveillance footage, scanned handwritten documents, and audio recordings of court hearings. Currently, our framework requires all non-textual evidence to be transcribed or described textually before processing, which may result in the loss of critical visual or auditory nuances essential for fact verification. For instance, distinguishing between ``inten'' and ``negligence'' might sometimes rely on visual cues in surveillance video that textual descriptions fail to capture fully. Extending the \textit{Hierarchical Legal Knowledge Graph} to incorporate multimodal nodes (e.g., embedding visual evidence into the \textit{Fact Graph}) represents a promising avenue for future research. Such an extension would enable the model to perform cross-modal reasoning, verifying textual testimony against visual evidence, thereby moving closer to a holistic and robust ``Smart Court'' system.

\section*{Ethics Statement}
We confirm that this study fully complies with the ACL Ethics Policy. Below, we address specific ethical considerations regarding the data and the application of our proposed model, LegalGraphRAG.

\paragraph{Data Privacy and Compliance} Our experiments involve four publicly available datasets (CAIL2018, CMDL, JuDGE, and LeCaRDv2) and statutory texts. These resources are established benchmarks in the legal NLP community. We emphasize that all court judgments utilized in this work have been pre-processed and anonymized by the original data providers. Private details, including the real names of defendants and victims, have been removed or masked to ensure no personally identifiable information (PII) is exposed. We strictly use this data for academic research purposes and adhere to their respective data usage licenses.

\paragraph{Bias and Fairness} We acknowledge that models trained on historical legal judgment data may inadvertently capture or amplify inherent biases present in the judicial system, such as those related to region or gender. While our work focuses on improving the logical reasoning and retrieval capabilities of legal LLMs through GraphRAG, where the outputs are interpreted with clear evidence.

\paragraph{Intended Use and Misuse} The proposed LegalGraphRAG is designed as an assistive tool to support legal professionals and researchers in retrieving precedents and analyzing case facts. It is \textbf{not} intended to replace human judges or lawyers, nor should it be deployed as a fully automated decision-making system in real-world judicial scenarios. The ``prison term'' and ``judgment'' predictions generated by the model should be viewed as reference probabilities rather than enforceable verdicts.
\bibliography{custom}

\clearpage
\appendix

\label{sec:appendix}
\newpage
\section{Frequently Asked Questions (FAQs)}

\subsection{What are the advantages of LegalGraphRAG?}

LegalGraphRAG introduces several key advancements over traditional retrieval-augmented generation methods and specialized legal LLMs, addressing critical challenges in the legal domain through its hierarchical structure and multi-agent workflow.

\paragraph{Superior Retrieval Effectiveness.}
First, our framework significantly improves how legal information is retrieved. While traditional flat retrieval methods often struggle to differentiate between specific case facts and abstract statutory rules, our hierarchical graph organizes this complex information into distinct levels. This structure ensures that the system captures both detailed evidence and high-level principles, providing a much more comprehensive context than standard baselines.

\paragraph{Trustworthy and Transparent Reasoning.} 
Second, LegalGraphRAG addresses the ``black box'' issue common in standard LLMs. Instead of generating answers directly, which can lead to hallucinations or correct predictions based on wrong premises, our system employs a multi-agent workflow. This process strictly verifies the retrieved evidence against the facts of the case. Consequently, it constructs a logical chain of evidence, ensuring that the final judgment is grounded in valid legal logic rather than statistical probability.

\paragraph{Flexibility and Model Agnosticism.} 
Finally, the framework offers superior flexibility compared to rigid, specialized legal models. Unlike methods that require extensive and costly fine-tuning on legal datasets, LegalGraphRAG functions as a modular system. It allows users to easily swap the underlying backbone model. As demonstrated in our experiments, this capability enables the integration of powerful advanced models to achieve state-of-the-art performance without the need for additional training.

\subsection{How to Evaluate Legal Reasoning?}
We utilize Legal Judgment Prediction (LJP) as the primary experimental testbed because it serves as a rigorous ``cognitive touchstone'' for evaluating complex reasoning in specialized domains. While our introduction highlights broader challenges in healthcare, finance, and law, LJP uniquely encapsulates the core difficulties of high-stakes reasoning.

\paragraph{Validating Hierarchical Knowledge Alignment} 
Introduction Challenge (i) highlights the difficulty of managing heterogeneous knowledge. LJP exemplifies this struggle by requiring the model to bridge the semantic gap between concrete case facts and abstract statutory rules. Successfully mapping these distinct granularities validates the effectiveness of our hierarchical graph structure in organizing multi-level domain knowledge.
    
\paragraph{Evaluating Rigorous Logical Deduction} 
Unlike general Question Answering tasks that may rely on surface-level semantic matching, LJP necessitates strict syllogistic reasoning (Major Premise $\rightarrow$ Minor Premise $\rightarrow$ Conclusion). This structural dependency provides an ideal setting to stress-test our Multi-Agent framework, specifically validating whether the \textit{Auditor} can effectively filter irrelevant distractions and enforce the logical consistency.
    
\paragraph{Benchmarking High-Stakes Reliability} 
In professional domains, plausibility is insufficient; accuracy is paramount. LJP imposes a zero-tolerance standard for hallucination, as every judgment must be supported by cited articles. By demonstrating that LegalGraphRAG can produce verifiable, evidence-based judgments in this demanding context, we establish a strong precedent for its applicability to critical domains like medicine and finance.

\subsection{How was the retrieval performance evaluated and compared across different strategies?}\label{sec:Retrieval Effectiveness}
To ensure consistent assessment throughout our study (spanning both the preliminary investigation and main comparative experiments), we established a standardized evaluation pipeline based on the legal corpora and CAIL~\cite{xiao2018cail2018} dataset described in Appendix~\ref{sec:Corpus}. The evaluation procedure consists of three steps.

\paragraph{Execution on Test Set}
For every case query in the test dataset, we executed two representative RAG strategies. The \textit{Flat Strategy} follows the traditional baseline approach, indexing all legal documents in a unified flat repository. The \textit{Hierarchical Strategy}, built upon Naive RAG, adopts a decoupled approach by separately storing and retrieving legal articles and historical cases. Based on these strategies, each model retrieved a set of candidate evidence from the corpus.
    
\paragraph{Ground Truth Alignment}
We utilized the articles provided in the dataset (ground truth articles) as the ``Gold Standard,'' as all cases in the dataset are inherently annotated with their relevant statutory articles. Any retrieved node matching these articles was marked as a \textit{True Positive}.
    
\paragraph{Metric Calculation} Based on the alignment results, we quantified performance using the two key indicators defined in Appendix~\ref{sec:Evaluation Metrics}:
\begin{itemize}
    \item \textit{Retrieval Effectiveness:} This measures the Recall of gold-standard evidence, indicating the system's ability to locate legal evidence.
    \item \textit{Error Rate:} This assesses the proportion of irrelevant or misleading nodes within the retrieved context reflecting the system's ability to filter distractions.
\end{itemize}

This allows us to objectively compare how different structural approaches (flat vs. hierarchical) impact the precision of legal reasoning.

\subsection{How were the ``context with irrelevant information'' constructed for the Generation Quality investigation?}\label{sec:context with irrelevant information}

To rigorously test the model's verification capabilities, we constructed evaluation contexts containing High-Similarity Irrelevant Information. Instead of including randomly selected texts, we curated sets of documents that are semantically similar to the correct evidence but legally inapplicable. This design mirrors real-world scenarios where documents share surface-level keywords but differ fundamentally in domain applicability. The construction process involved two steps:

\paragraph{Ground Truth Context:} 
First, we established the baseline context using the ground truth provided in the CAIL~\cite{xiao2018cail2018} dataset. For each case, this set consists exclusively of the correct applicable articles required for the judgment.
    
\paragraph{Injection of Irrelevant Distractors:} 
To simulate the presence of legally plausible but factually irrelevant documents, we utilized the entire Criminal Law code as a retrieval corpus. For each correct article in the Ground Truth Context, we performed a vector-based similarity search over this corpus to identify the top-$k$ most similar articles that were not part of the ground truth. 

These retrieved articles serve as High-Similarity Distractors: they share significant lexical and semantic overlap with the correct laws (e.g., sharing keywords like ``theft'' or ``fraud'') but differ in specific constitutive elements or sentencing standards. By mixing these irrelevant documents into the context, we created a challenging environment that forces the model to discern legal essence from superficial similarity.

\subsection{Reliability Analysis Definitions}
\label{sec:reliability_detail}
We analyzed the CAIL test results to categorize correct predictions based on evidence support. A prediction is classified as Traceable Correct if the model correctly predicts the charge and successfully retrieves the ground-truth articles. Conversely, it is Untraceable Correct if the correct charge is predicted despite failing to retrieve the necessary articles.

\begin{figure*}[t]
    \centering
    \includegraphics[width=0.95\linewidth]{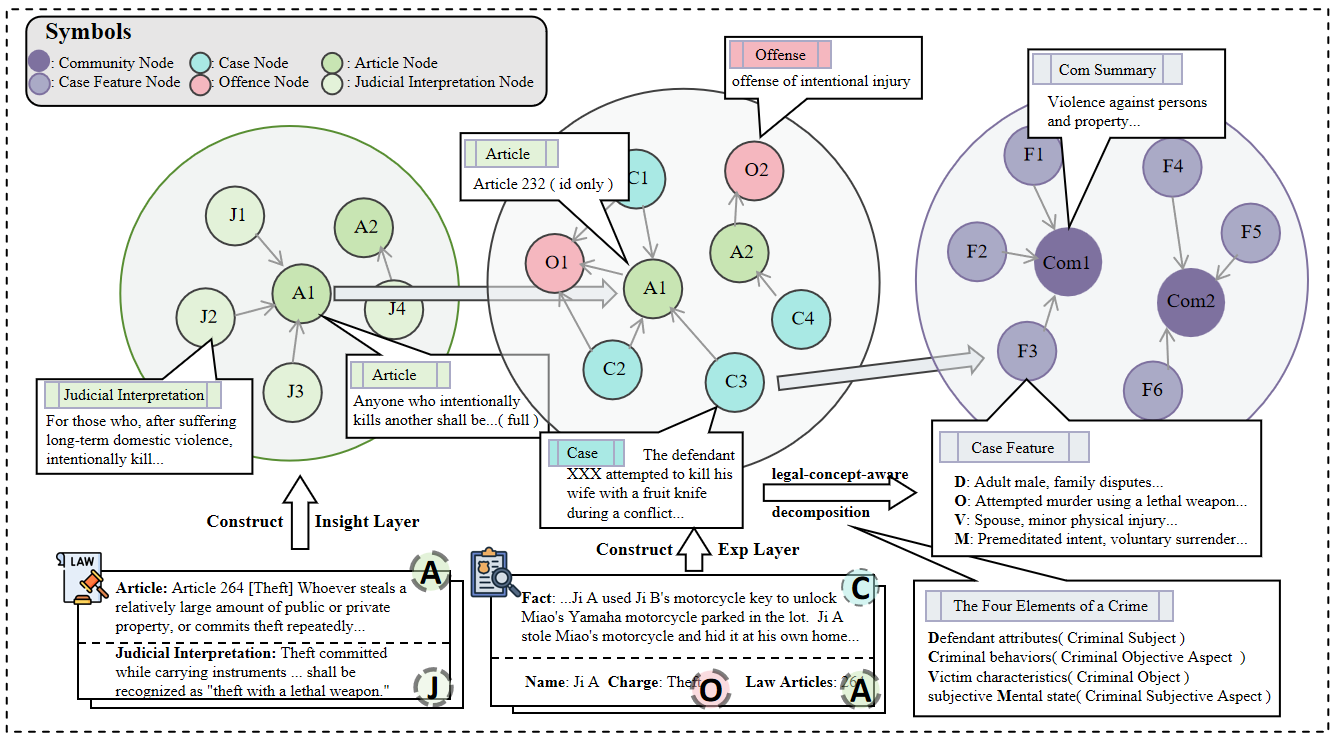}
    \caption{Overview of the Hierarchical Knowledge Construction phase in LegalGraphRAG.}
    \label{fig:construct}
    \vspace{-2mm}
\end{figure*}

\begin{figure*}[t]
    \centering
    \includegraphics[width=0.95\linewidth]{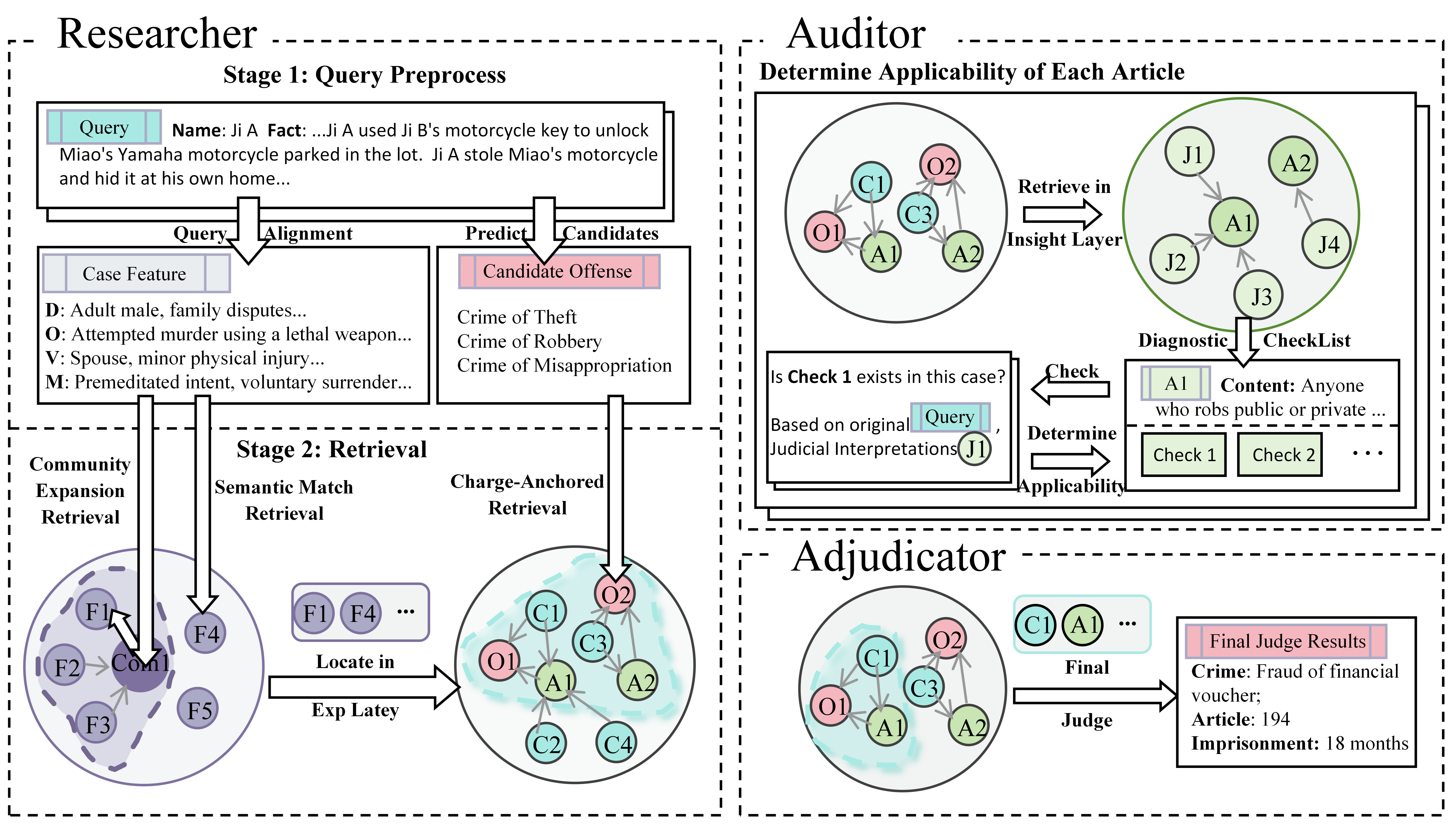}
    \caption{The workflow of the Evidence-based Legal Reasoning phase.}
    \label{fig:retrieval}
    \vspace{-2mm}
\end{figure*}

\section{Method Details}\label{sec: Method Details}
In this section, we provide the comprehensive technical specifications and implementation details of the proposed \textbf{LegalGraphRAG} framework. As outlined in the main text, our approach operates in two distinct phases: (i) \textbf{Hierarchical Knowledge Construction} (Figure~\ref{fig:construct}), which organizes legal knowledge into a layered graph structure to effectively decouple historical cases, relevant articles, and judicial interpretations; and (ii) \textbf{Evidence-based Legal Reasoning} (Figure~\ref{fig:retrieval}), which employs a collaborative agent workflow to retrieve relevant evidence and generate verifiable judgments.

\subsection{Hierarchical Knowledge Construction}\label{sec:construct detail}
We construct a Hierarchical Legal Graph $\mathcal{G}$ composed of three specialized subgraphs, as illustrated in Figure~\ref{fig:graph_structure}. This multi-layered structure explicitly differentiates between specific precedents, abstract case relationships, and rigorous statutory rules, as illustrated in Figure \ref{fig:construct}.

\noindent \textbf{Fact Graph ($\mathcal{G}_{fac}$)} serves as the repository for ground-truth precedents. It encodes the natural structure of legal documents by explicitly linking Cases ($\mathcal{C}$), Articles ($\mathcal{A}$), and Offenses ($\mathcal{O}$). Edges are established to represent citation relationships ($e_{ca}: \mathcal{C} \rightarrow \mathcal{A}$) and conviction outcomes ($e_{co}: \mathcal{C} \rightarrow \mathcal{O}$). Formally, it is defined as:
\begin{equation}
\mathcal{G}_{fac}
= (\mathcal{V}_{fac}, \mathcal{E}_{fac})
= \left(
    \bigl\{ c_i, a_i, o_i \bigr\}_{i=1}^{|\mathcal{G}_{fac}|}\right).
\end{equation}

\noindent \textbf{Ontology Graph ($\mathcal{G}_{ont}$)} abstracts case features to model inter-case relationships. To map unstructured narratives into a structured semantic space, we define a domain-specific ontology along four dimensions: \textit{Defendant Attributes}, \textit{Criminal Behaviors}, \textit{Victim Characteristics}, and \textit{Subjective Mental States}. Keywords are extracted and aligned with these dimensions to form Case Feature Nodes ($\mathcal{F}$). 

Structurally, we utilize the k-Nearest Neighbors (k-NN) algorithm to establish semantic edges between cases. Based on this topology, we apply the Leiden algorithm~\cite{traag2019louvain} to cluster related cases into Community Nodes ($\mathcal{K}$), facilitating coarse-to-fine retrieval. The subgraph is formally defined as:
\begin{equation}
\mathcal{G}_{ont}
= (\mathcal{V}_{ont}, \mathcal{E}_{ont})
= \left(
    \bigl\{ c_i, k_j \bigr\}_{i=1,j=1}^{|\mathcal{G}_{ont}|}
\right) .
\end{equation}

\noindent \textbf{Rule Graph ($\mathcal{G}_{rul}$)} incorporates fine-grained legal knowledge to resolve statutory ambiguities. This graph consists of Articles ($\mathcal{A}$) and Judicial Interpretations ($\mathcal{J}$), linked by explicit cross-references. To further enhance precision, each article node $a_i$ is equipped with a \textbf{Diagnostic Checklist} $\mathcal{D}(a_i)$. 

Generated by parsing statutory texts, this checklist decomposes complex legal provisions into atomic boolean queries. For instance, regarding \textit{Article 266} (Fraud), the checklist validates the logical chain of the crime: ``\textit{Did the defendant fabricate facts or conceal the truth?}'', ``\textit{Did the victim fall into a mistake due to this act?}'', and ``\textit{Did the victim dispose of property based on this mistake?}''. This mechanism forces the model to verify each constitutive element step-by-step, rather than relying on vague semantic overlaps. Formally, this subgraph and its associated checklists are defined as:
\begin{equation}
\mathcal{G}_{rul}
= (\mathcal{V}_{rul}, \mathcal{E}_{rul})
= \left(
\bigl\{ a_i, j_i \bigr\}_{i=1}^{|\mathcal{G}_{rul}|} \right),
\end{equation}
where
\begin{equation}
\mathcal{D}(a_i) = \{d_1,\dots,d_{|C|}\}.
\end{equation}

By integrating these three layers, HierarGraph $\mathcal{G}$ transforms heterogeneous legal corpora into a structured ecosystem. This architecture directly addresses the limitations of flat retrieval by offering multi-granular support for our multi-agent system.

\begin{algorithm*}[t]
\caption{Evidence-based Legal Reasoning}
\label{alg:legal_reasoning}
\begin{algorithmic}[1]
\Require 
    Raw case query $q$; 
    Ontology Graph $\mathcal{G}_{ont}$; 
    Fact Graph $\mathcal{G}_{fac}$; 
    Rule Graph $\mathcal{G}_{rul}$.
\Ensure 
    Final Judgment $\mathcal{J}$ with citations.

\Statex \textbf{Stage 1: Researcher Agent (Multi-Strategy Retrieval)}
\State $\phi(q) \leftarrow \text{OntologyAlign}(q, \mathcal{G}_{ont})$ \Comment{Align query to ontology features}

\Statex \textit{Parallel Evidence Retrieval Strategies:}
\State $\mathcal{S}_{cand} \leftarrow \mathcal{R}_{\text{sem}} \cup \mathcal{R}_{\text{com}} \cup \mathcal{R}_{\text{chg}}$ \Comment{Union of candidate evidence}

\Statex \textbf{Stage 2: Auditor Agent (Verification \& Pruning)}
\State $\mathcal{S}_{verified} \leftarrow \mathcal{S}_{cand}$ 
\For{each article node $v_a \in \mathcal{S}_{cand}$}
    \State $\mathcal{D} \leftarrow \text{RetrieveChecklist}(v_a, \mathcal{G}_{rul})$ \Comment{Get checklist $\mathcal{D}(v_a)=\{d_1,\dots,d_{|C|}\}$}
    \State $\mathcal{J} \leftarrow \text{RetrieveInterpretations}(v_a, \mathcal{G}_{rul})$ \Comment{Get Judicial Interpretations}
    
    \State $V_{results} \leftarrow \emptyset$
    \For{each diagnostic item $d_k \in \mathcal{D}$} \Comment{Item-wise verification loop}
        \State $r_k \leftarrow \text{CheckCondition}(q, d_k, v_a, \mathcal{J})$ \Comment{Verify if fact $q$ satisfies condition $d_k$}
        \State $V_{results} \leftarrow V_{results} \cup \{r_k\}$
    \EndFor
    
    \State $IsApplicable \leftarrow \text{Decide}(V_{results})$ \Comment{Final determination for node $v_a$}
    
    \If{ not $IsApplicable$}
        \State $\mathcal{S}_{verified} \leftarrow \text{Prune}(\mathcal{S}_{verified}, v_a)$ \Comment{Remove article and linked nodes}
    \EndIf
\EndFor

\State $\mathcal{G}_{sub}^f \leftarrow \{V_A^f, V_C^f, V_O^f\} \leftarrow \text{Organize}(\mathcal{S}_{verified})$ \Comment{Structure the verified subgraph}

\Statex \textbf{Stage 3: Adjudicator Agent (Synthesis)}
\State $\mathcal{Y} \leftarrow \text{Adjudicator}(q \oplus V_A^f \oplus V_C^f \oplus V_O^f)$ \Comment{Synthesize judgment with citations}
\State \Return $\mathcal{Y}$
\end{algorithmic}
\end{algorithm*}

\subsection{Evidence-based Legal Reasoning}
We propose a multi-agent framework to emulate the rigorous workflow of legal professionals, as illustrated in Figure \ref{fig:retrieval}. This system operates sequentially through three specialized agents (\textbf{Researcher}, \textbf{Auditor}, and \textbf{Adjudicator}) to transform a case query into a verifiable judgment.

\vspace{1mm}
\noindent \textbf{Researcher Agent}\label{sec:researcher detail}: This agent is responsible for grounding the unstructured case query in relevant legal knowledge. First, it aligns the raw case description with the ontology structure in $\mathcal{G}_{ont}$, extracting standardized evidentiary features (e.g., defendant characteristics and criminal behaviors). 

Based on these features, we formulate the evidence retrieval process $\mathcal{R}(q)$ as the union of three parallel strategies, where $q$ is the legal query:
\begin{equation}
\mathcal{R}(q) = \mathcal{R}_{\text{sem}}(q) \cup \mathcal{R}_{\text{com}}(q) \cup \mathcal{R}_{\text{chg}}(q)
\end{equation}

\noindent (i) \textbf{Semantic Match Retrieval}: We first locate direct evidentiary analogues via fine-grained semantic similarity. Let $\phi(\cdot)$ denote the ontology-aligned embeddings, this process is defined as:
\begin{equation}
\mathcal{R}_{\text{sem}}(q) = \operatorname*{Top-k}_{c \in \mathcal{G}_{ont}} \text{sim}\bigl(\phi(q), \phi(c)\bigr)
\end{equation}

\noindent (ii) \textbf{Community Expansion Retrieval}: To capture broader structural context, we employ a community-guided strategy. We identify the single most relevant thematic community $\mathcal{K}^*$ aligned with the query, and then retrieve the top-$k$ similar cases restricted within this community:
\begin{equation}
\begin{split}
    \mathcal{K}^* &= \operatorname*{argmax}_{\mathcal{K} \in \mathcal{G}_{ont}} \text{sim}\bigl(\phi(q), \phi(\mathcal{K})\bigr) \\
    \mathcal{R}_{\text{com}}(q) &= \operatorname*{Top-k}_{c \in \mathcal{K}^*} \text{sim}\bigl(\phi(q), \phi(c)\bigr)
\end{split}
\end{equation}

\noindent (iii) \textbf{Charge-Anchored Retrieval}: Finally, we anchor the legal basis by retrieving cases linked to inferred charges. Here, $\mathcal{O}(q)$ denotes the set of predicted charges and $\mathcal{N}_{\mathcal{G}_{fac}}(o)$ represents the neighboring cases connected to charge $o$ in the Fact Graph:
\begin{equation}
\mathcal{R}_{\text{chg}}(q) = \bigcup_{o \in \mathcal{O}(q)} \mathcal{N}_{\mathcal{G}_{fac}}(o)
\end{equation}

These three retrieval strategies organize the candidate evidence set $\mathcal{S}_{cand}$.

\vspace{1mm}
\noindent \textbf{Auditor Agent}\label{sec:auditor detail}: Operating on the candidate evidence set $\mathcal{S}_{cand}$, the Auditor validates the applicability of each retrieved article $v_a$ through a rigorous verify-and-prune mechanism. This process proceeds in three specific steps:

\noindent (i) \textbf{Diagnostic Retrieval}: For each article $v_a$, the agent retrieves its specific Diagnostic Checklist $\mathcal{D}(v_a)=\{d_1,\dots,d_{|C|}\}$ and relevant Judicial Interpretations $\mathcal{J}$ from the Rule Graph $\mathcal{G}_{rul}$.

\noindent (ii) \textbf{Item-wise Verification}: The agent executes a verification loop for each diagnostic item $d_k \in \mathcal{D}$. It evaluates whether the raw case facts $q$ satisfy the specific legal condition $d_k$, supporting the judgment with the interpretive context $\mathcal{J}$. This produces a set of boolean verification results $V_{results} = \{r_k\}$, where $r_k \leftarrow \text{CheckCondition}(q, d_k, v_a, \mathcal{J})$.

\noindent (iii) \textbf{Decision and Pruning}: Finally, the Auditor synthesizes the verification results $V_{results}$ to determine the overall applicability of the article. If the article fails to meet the necessary criteria ($IsApplicable$ is False), the Auditor executes a pruning operation:
\begin{equation}
\mathcal{S}_{verified} \leftarrow \text{Prune}(\mathcal{S}_{verified}, v_a)
\end{equation}
This step removes the inapplicable article node $v_a$ along with its dependent case precedents and charge nodes, ensuring that the final subgraph $\mathcal{S}_{verified}$ contains only logically valid and applicable evidence.

\vspace{1mm}
\noindent \textbf{Adjudicator Agent}: The Adjudicator synthesizes the verified subgraph $\mathcal{S}_{verified}$ to render the final judgment. Specifically, it organizes the valid nodes extracted from $\mathcal{S}_{verified}$ into sets of confirmed articles ($V_A^f$), case precedents ($V_C^f$), and charge information ($V_O^f$). By integrating these evidence components with the original query $q$, it generates a response with explicit citations. This process is formulated as:
\begin{equation}
    \mathcal{Y} = Adjudicator (q \oplus V_A^f \oplus V_C^f \oplus V_O^f)
\end{equation}
The output $\mathcal{Y}$ ensures that every conclusion is directly traceable to specific nodes in the knowledge graph, enforcing transparency and evidence-based reasoning.

\begin{table*}[t]
\centering
\resizebox{\textwidth}{!}{
\begin{tabular}{ll|cc|cc|cc|cc|cc}
\toprule
& & \multicolumn{10}{c}{\cellcolor{headerblue}\textbf{CAIL}} \\
\cmidrule{3-12}
\multirow{2}{*}{\textbf{Model}} & \multirow{2}{*}{\textbf{Size}} & \multicolumn{2}{c|}{\textbf{Public Safety}} & \multicolumn{2}{c|}{\textbf{Economic}} & \multicolumn{2}{c|}{\textbf{Social Order}} & \multicolumn{2}{c|}{\textbf{Person Rights}} & \multirow{2}{*}{\textbf{All}} & \multirow{2}{*}{$\Delta$} \\
\cmidrule(lr){3-4} \cmidrule(lr){5-6} \cmidrule(lr){7-8} \cmidrule(lr){9-10}
& & \textbf{ACC} & \textbf{F1} & \textbf{ACC} & \textbf{F1} & \textbf{ACC} & \textbf{F1} & \textbf{ACC} & \textbf{F1} & & \\
\midrule
\multicolumn{12}{c}{\textbf{GPT-4o-mini}} \\
\midrule
Naive RAG & $\sim$8B & 27.5& 37.6& 18.8& 33.6& 18.0& 28.8& 22.1& 39.2& 22.2& \cellcolor{midgreen}$\uparrow$ 18.7 \\
G-Retriever & $\sim$8B & 17.5& 24.8& 20.3& 32.8& 20.5& 31.0& 24.1& 31.6& 21.4& \cellcolor{midgreen}$\uparrow$ 19.5 \\
LightRAG & $\sim$8B & 25.4& 37.4& 21.3& 36.9& 21.7& 38.7& 23.4& 42.8& 23.1& \cellcolor{midgreen}$\uparrow$ 17.8 \\
RAPTOR~\cite{sarthi2024raptor} & $\sim$8B & \underline{33.1}& 49.0& \underline{29.3}& 44.2& \underline{25.9}& 39.9& 28.3& 43.1& 30.5& \cellcolor{lightgreen}$\uparrow$ 10.4 \\
HippoRAG2~\cite{gutierrez2025rag} & $\sim$8B & \underline{33.1}& \underline{50.1}& 28.1& \underline{46.1}& 21.7& \underline{43.6}& \underline{37.2}& \underline{54.2}& \underline{31.9}& \cellcolor{lightgreen}$\uparrow$ 9.0 \\
\textbf{LegalGraphRAG}~(Ours) & $\sim$8B & \textbf{39.6}& \textbf{54.8}& \textbf{36.3}& \textbf{52.9}& \textbf{37.3}& \textbf{51.2}& \textbf{42.1}& \textbf{62.4}& \textbf{40.9}& -- \\
\midrule
\multicolumn{12}{c}{\textbf{DeepSeek-V3.1}} \\
\midrule
Naive RAG & $\sim$200B & 38.0& 54.0& 32.3& 49.9& 33.4& 47.3& 40.7& \underline{53.4}& 37.8& \cellcolor{midgreen}$\uparrow$ 12.1 \\
G-Retriever & $\sim$200B & 36.5& 54.6& 35.1& 49.8& 36.2& 47.2& 39.5& 48.3& 37.2& \cellcolor{midgreen}$\uparrow$ 12.7 \\
LightRAG & $\sim$200B & 36.6& 48.5& 26.3& 50.2& 33.5& 46.3& 39.7& 53.1& \underline{45.4}& \cellcolor{lightgreen}$\uparrow$ 4.5 \\
RAPTOR~\cite{sarthi2024raptor} & $\sim$200B & \underline{42.2}& \underline{56.3}& \underline{37.8}& \underline{53.0}& \underline{39.2}& \underline{50.7}& \underline{45.5}& 52.3& 44.4& \cellcolor{lightgreen}$\uparrow$ 5.5 \\
HippoRAG2~\cite{gutierrez2025rag} & $\sim$200B & 41.5& 49.1& 33.1& 46.8& 34.3& 47.0& 38.6& 46.4& 41.2& \cellcolor{lightgreen}$\uparrow$ 8.7 \\
\textbf{LegalGraphRAG}~(Ours) & $\sim$200B & \textbf{44.4}& \textbf{58.8}& \textbf{41.9}& \textbf{57.8}& \textbf{41.9}& \textbf{56.8}& \textbf{46.2}& \textbf{65.1}& \textbf{49.9}& -- \\
\bottomrule
\end{tabular}
}
\label{tab:more_results_closed_models_CAIL}

\centering
\resizebox{\textwidth}{!}{
\begin{tabular}{ll|cc|cc|cc|cc|cc}
\toprule
& & \multicolumn{10}{c}{\cellcolor{headerpurple}\textbf{CMDL}} \\
\cmidrule{3-12}
\multirow{2}{*}{\textbf{Model}} & \multirow{2}{*}{\textbf{Size}} & \multicolumn{2}{c|}{\textbf{Public Safety}} & \multicolumn{2}{c|}{\textbf{Economic}} & \multicolumn{2}{c|}{\textbf{Social Order}} & \multicolumn{2}{c|}{\textbf{Person Rights}} & \multirow{2}{*}{\textbf{All}} & \multirow{2}{*}{$\Delta$} \\
\cmidrule(lr){3-4} \cmidrule(lr){5-6} \cmidrule(lr){7-8} \cmidrule(lr){9-10}
& & \textbf{ACC} & \textbf{F1} & \textbf{ACC} & \textbf{F1} & \textbf{ACC} & \textbf{F1} & \textbf{ACC} & \textbf{F1} & & \\
\midrule
\multicolumn{12}{c}{\textbf{GPT-4o-mini}} \\
\midrule
Naive RAG & $\sim$8B & 38.7& 50.8& 35.6& 44.0& 30.9& 44.2& 33.3& 43.0& 32.9& \cellcolor{lightgreen}$\uparrow$ 17.1 \\
G-Retriever & $\sim$8B & 24.6& 38.4& 29.7& 38.8& 30.4& 40.0& 41.1& 51.6& 28.3& \cellcolor{midgreen}$\uparrow$ 21.7 \\
LightRAG & $\sim$8B & 36.2& 43.1& 37.5& 48.9& 46.9& 55.1& 34.6& 50.8& 34.2& \cellcolor{lightgreen}$\uparrow$ 15.8 \\
RAPTOR~\cite{sarthi2024raptor} & $\sim$8B & 42.0& 49.1& \underline{48.2}& 58.0& \underline{57.3}& 60.9& 46.0& 59.5& \underline{48.7}& \cellcolor{lightgreen}$\uparrow$ 11.3 \\
HippoRAG2~\cite{gutierrez2025rag} & $\sim$8B & \underline{45.0}& \underline{52.5}& 42.9& \underline{60.8}& 45.3& \underline{64.6}& \underline{59.4}& \underline{75.0}& 46.0& \cellcolor{lightgreen}$\uparrow$ 14.0 \\
\textbf{LegalGraphRAG}~(Ours) & $\sim$8B & \textbf{46.5}& \textbf{57.8}& \textbf{63.7}& \textbf{67.9}& \textbf{58.0}& \textbf{66.2}& \textbf{67.2}& \textbf{75.4}& \textbf{60.0}& -- \\
\midrule
\multicolumn{12}{c}{\textbf{DeepSeek-V3.1}} \\
\midrule
Naive RAG & $\sim$200B & 52.0& 65.3& 66.1& 71.6& 69.8& 70.7& 68.8& \underline{80.5}& 62.9& \cellcolor{lightgreen}$\uparrow$ 12.8 \\
G-Retriever & $\sim$200B & 46.2& \underline{64.9}& 58.6& 69.6& 56.1& 67.4& 69.7& 78.9& 55.2& \cellcolor{midgreen}$\uparrow$ 23.5 \\
LightRAG & $\sim$200B & 47.7& 58.7& 46.5& 67.6& 47.6& 53.2& 52.3& 64.2& 52.4& \cellcolor{midgreen}$\uparrow$ 26.3 \\
RAPTOR~\cite{sarthi2024raptor} & $\sim$200B & 53.4& 64.7& \underline{68.2}& \underline{73.0}& 66.4& 75.4& 60.3& 71.1& 56.4& \cellcolor{lightgreen}$\uparrow$ 12.3 \\
HippoRAG2~\cite{gutierrez2025rag} & $\sim$200B & \underline{62.1}& 63.5& 62.4& 65.3& \textbf{75.9}& \underline{78.6}& \underline{76.9}& 78.2& \underline{74.0}& \cellcolor{lightgreen}$\uparrow$ 4.7 \\
\textbf{LegalGraphRAG}~(Ours) & $\sim$200B & \textbf{66.7}& \textbf{69.9}& \textbf{76.0}& \textbf{79.3}& \underline{72.9}& \textbf{80.4}& \textbf{79.7}& \textbf{85.5}& \textbf{78.7}& -- \\
\bottomrule
\end{tabular}
}
\caption{\textbf{Performance comparison on advanced Models}. We compared LegalGraphRAG and other baselines utilizing advanced LLMs as backbones.}
\label{tab:more_results_closed_models_CMDL}
\end{table*}

\begin{table*}[p]
\centering
\resizebox{\textwidth}{!}{
\begin{tabular}{ll|cc|cc|cc|cc|cc}
\toprule
& & \multicolumn{10}{c}{\cellcolor{headerblue}\textbf{CAIL}} \\
\cmidrule{3-12}
\multirow{2}{*}{\textbf{Model}} & \multirow{2}{*}{\textbf{Size}} & \multicolumn{2}{c|}{\textbf{Public Safety}} & \multicolumn{2}{c|}{\textbf{Economic}} & \multicolumn{2}{c|}{\textbf{Social Order}} & \multicolumn{2}{c|}{\textbf{Person Rights}} & \multirow{2}{*}{\textbf{All}} & \multirow{2}{*}{$\Delta$} \\
\cmidrule(lr){3-4} \cmidrule(lr){5-6} \cmidrule(lr){7-8} \cmidrule(lr){9-10}
& & \textbf{ACC} & \textbf{F1} & \textbf{ACC} & \textbf{F1} & \textbf{ACC} & \textbf{F1} & \textbf{ACC} & \textbf{F1} & & \\
\midrule
\multicolumn{12}{c}{\textbf{Open-Source Models}} \\
\midrule
Qwen-2.5-7B-Instruct & 7B-Inst & 28.7& 53.8& 24.1& 48.2& 26.6& 53.5& 36.0& 56.2& 30.1&\cellcolor{darkgreen}$\uparrow$ 17.8 \\
Qwen-3-8B & 8B-Inst & 23.9& 58.5& 27.6& 51.2& 36.6& 62.2& 46.2& 66.7& 35.9&\cellcolor{midgreen}$\uparrow$ 12.0 \\
Internlm3-8b-instruct & 8B-Inst & 27.6& 59.9& 26.4& 52.8& 30.8& 59.0& 33.6& 57.6& 29.9&  \cellcolor{midgreen}$\uparrow$ 18.0 \\
Glm-4-9b-chat & 9B-Inst & 23.9& 60.1& 25.6& 51.0& 34.7& 59.4& 35.1& 58.5& 30.8& \cellcolor{midgreen}$\uparrow$ 17.1 \\
\midrule
\multicolumn{12}{c}{\textbf{Advanced Models}} \\
\midrule
GPT-4o-mini~\cite{achiam2023gpt} & $\sim$8B & 24.7& 57.2 & 25.7& 42.4& 24.6& 51.8& 34.7& 55.0& 30.9& \cellcolor{midgreen}$\uparrow$ 17.0 \\
DeepSeek-V3.1~\cite{liu2024deepseek} & $\sim$200B & \underline{42.3}& \underline{63.2}& \underline{37.1}& \textbf{61.3}& \underline{44.5}& \underline{68.8}& \underline{51.9}& \underline{67.3}& \underline{44.9}& \cellcolor{lightgreen}$\uparrow$ 3.0 \\
\midrule
\multicolumn{12}{c}{\textbf{Legal Specific Methods}} \\
\midrule
DISC-LawLLM-7B~\cite{yue2024lawllm} & 7B-Inst & 36.5& 55.9& 29.9& 48.8& 41.5& 61.2& 39.3& 57.8& 38.2& \cellcolor{darkgreen}$\uparrow$ 9.7 \\
ADAPT~\cite{deng2024enabling} & 7B-Inst & 40.1 & 50.8& 31.6& 42.4& 39.6& 54.9& 41.0& 50.5& 41.3& \cellcolor{lightgreen}$\uparrow$ 6.6 \\
Legal $\Delta$~\cite{dai2025legal} & 7B-Inst & 33.0& 54.9& 27.8& 51.0& 34.6& 59.4& 44.5& 60.6& 37.9& \cellcolor{lightgreen}$\uparrow$ 10.0 \\
\midrule
\multicolumn{12}{c}{\textbf{RAG Based Methods}} \\
\midrule
Naive RAG & 8B-Inst & 30.4& 45.7& 29.2& 48.6& 37.5& 54.1& 36.6& 51.1& 34.8& \cellcolor{midgreen}$\uparrow$ 13.1 \\
RAPTOR~\cite{sarthi2024raptor} & 8B-Inst & 36.4& 59.2& 33.4& 53.9& 38.9& 64.1& 41.0& 61.7& 37.2&\cellcolor{lightgreen}$\uparrow$ 10.7 \\
HippoRAG2~\cite{gutierrez2025rag} & 8B-Inst & 35.2& 60.3& 33.1& 53.7& 41.7& 67.0& 42.6& 63.8& 39.8& \cellcolor{lightgreen}$\uparrow$ 8.1 \\
\textbf{LegalGraphRAG}~(Ours) & 8B-Inst &\textbf{43.0}& \textbf{64.9}& \textbf{37.8}& \underline{61.0}& \textbf{44.6}& \textbf{69.4}& \textbf{54.5}& \textbf{70.6}& \textbf{47.9}& -- \\
\bottomrule
\end{tabular}
}
\caption{\textbf{Extended experiments on Article Prediction}. We evaluated the performance of our model and baselines on the specific sub-task of law article prediction. We visualize the gains of LegalGraphRAG to the each baseline in the \colorbox{lightgreen}{$\Delta$ columns}.}
\label{tab:more_results_laws}

\centering
\resizebox{\textwidth}{!}{
\begin{tabular}{ll|cc|cc|cc|cc|cc}
\toprule
& & \multicolumn{10}{c}{\cellcolor{headerblue}\textbf{CAIL}} \\
\cmidrule{3-12}
\multirow{2}{*}{\textbf{Model}} & \multirow{2}{*}{\textbf{Size}} & \multicolumn{2}{c|}{\textbf{Public Safety}} & \multicolumn{2}{c|}{\textbf{Economic}} & \multicolumn{2}{c|}{\textbf{Social Order}} & \multicolumn{2}{c|}{\textbf{Person Rights}} & \multirow{2}{*}{\textbf{All}} & \multirow{2}{*}{$\Delta$} \\
\cmidrule(lr){3-4} \cmidrule(lr){5-6} \cmidrule(lr){7-8} \cmidrule(lr){9-10}
& & \textbf{ACC} & \textbf{MAE} & \textbf{ACC} & \textbf{MAE} & \textbf{ACC} & \textbf{MAE} & \textbf{ACC} & \textbf{MAE} & & \\
\midrule
\multicolumn{12}{c}{\textbf{Open-Source Models}} \\
\midrule
Qwen-2.5-7B-Instruct & 7B-Inst & 13.0& 23.7& 8.1& 33.3& 6.0& 30.3& 9.7& 29.4& 29.5& \cellcolor{lightgreen}$\uparrow$ 8.4 \\
Qwen-3-8B & 8B-Inst & 8.3& 32.6& 11.3& 31.6& 17.9& 29.7& 11.4& 26.4& 27.5& \cellcolor{lightgreen}$\uparrow$ 7.4 \\
Internlm3-8b-instruct & 8B-Inst & 7.1& 35.2& 5.9& 37.2& 6.0& 32.5& 7.2& 37.3& 33.7&  \cellcolor{midgreen}$\uparrow$ 13.6 \\
Glm-4-9b-chat & 9B-Inst & 3.6& 36.0& 3.2& 32.9& 1.5& 35.1& 6.8& 38.6& 33.1& \cellcolor{midgreen}$\uparrow$ 13.0 \\
\midrule
\multicolumn{12}{c}{\textbf{Advanced Models}} \\
\midrule
GPT-4o-mini~\cite{achiam2023gpt} & $\sim$8B & 6.9& 38.2& 7.3& 34.2& 8.3& 31.7& 8.0& 34.6& 33.6& \cellcolor{midgreen}$\uparrow$ 13.5 \\
DeepSeek-V3.1~\cite{liu2024deepseek} & $\sim$200B & 7.1& 31.2& 8.1& 31.5& 10.4& 25.9& 8.4& 29.1& 29.1& \cellcolor{lightgreen}$\uparrow$ 8.6 \\
\midrule
\multicolumn{12}{c}{\textbf{Legal Specific Methods}} \\
\midrule
DISC-LawLLM-7B~\cite{yue2024lawllm} & 7B-Inst & 15.5& 24.5& 5.0& 33.2& 3.0& 37.9& 8.4& 34.5& 31.6& \cellcolor{midgreen}$\uparrow$ 11.5 \\
ADAPT~\cite{deng2024enabling} & 7B-Inst & 8.3& 21.8& 10.9& \underline{21.9}& 3.0& 24.3& 9.7& \underline{22.3}& \underline{20.4}& \cellcolor{lightgreen}$\uparrow$ 0.3 \\
Legal $\Delta$~\cite{dai2025legal} & 7B-Inst & 11.9& 25.0& 9.0& 29.0& 9.0& 27.5& 8.9& 27.1& 26.3& \cellcolor{lightgreen}$\uparrow$ 6.2 \\
\midrule
\multicolumn{12}{c}{\textbf{RAG Based Methods}} \\
\midrule
Naive RAG & 8B-Inst & 10.5& 29.8& 11.4& 30.8& \underline{18.1}& 21.7& 12.6& 24.3& 26.5& \cellcolor{lightgreen}$\uparrow$ 4.4 \\
RAPTOR~\cite{sarthi2024raptor} & 8B-Inst & 12.0& \underline{21.7}& 11.7& 28.3& 15.8& 26.1& \underline{16.2}& \textbf{21.8}& 24.3& \cellcolor{lightgreen}$\uparrow$ 4.2 \\
HippoRAG2~\cite{gutierrez2025rag} & 8B-Inst & \underline{13.1}& 23.0& \underline{12.7}& 25.8& 17.9& \underline{23.8}& 13.5& 23.4& 23.8& \cellcolor{lightgreen}$\uparrow$ 3.7 \\
\textbf{LegalGraphRAG}~(Ours) & 8B-Inst &\textbf{14.0}& \textbf{20.9}& \textbf{13.7}& \textbf{22.1}& \textbf{19.4}& \textbf{23.6}& \textbf{17.1}& 22.7& \textbf{20.1}& -- \\
\bottomrule
\end{tabular}
}
\caption{\textbf{Extended experiments on Term of Penalty Prediction}. We assessed the accuracy and error rates of imprisonment term predictions compared to baselines. We visualize the gains of LegalGraphRAG to the each baseline in the \colorbox{lightgreen}{$\Delta$ columns}.}
\label{tab:more_results_terms}
\end{table*}

\begin{table*}[t]
\centering
\footnotesize
\begin{tabular}{@{}lcccccc@{}}
\toprule
\multirow{2}{*}{Method} & 
\multirow{2}{*}{Indexing Time (s)} & 
\multicolumn{2}{c}{Token Consumption ($<10^6$)} & 
\multicolumn{2}{c}{Avg Cost} \\
\cmidrule(lr){3-6}
& & Prompt & Completion & Time & Token \\
\midrule
RAPTOR &  13696.90 & 5.64& 0.72& 5.86& 3589\\
HippoRAG2 &  4581.60 & 10.58& 2.79& 11.2& 5199\\
\textbf{LegalGraphRAG (Ours)} & 3687.49 & 3.97 & 0.78 & 46.1& 10664\\
\bottomrule
\end{tabular}
\caption{Comparison of Computational Efficiency: Offline Indexing vs. Online Inference. We report the total time and token usage for graph construction (Indexing) and the average cost per query (Online).}
\label{tab:cost}
\end{table*}

\section{Additional Experiments}\label{sec:additional_exp}

\subsection{Extensions to the Main Experiment (Q4)}
In this section, we conduct a series of extended experiments to verify the universality of our framework across different model architectures and its robustness in specific, high-difficulty legal sub-tasks.

\noindent\textbf{Obs.7. Universality across Advanced Backbones.}
To verify the universality of our framework, we extended the evaluation to advanced large language models, specifically DeepSeek-V3.1 and GPT-4o-mini. As shown in Table \ref{tab:more_results_closed_models_CMDL}, LegalGraphRAG consistently outperforms all baselines across both CAIL and CMDL datasets, regardless of the backbone model employed. Notably, even with the lighter GPT-4o-mini on the CMDL dataset, our method achieves a remarkable performance gain (e.g., significantly exceeding the strong baseline RAPTOR in Accuracy), while maintaining its lead with the more powerful DeepSeek-V3.1. This demonstrates that LegalGraphRAG’s structured reasoning capabilities effectively complement the generation power of various state-of-the-art LLMs, enhancing their precision in complex legal application scenarios independent of the underlying model architecture.

\noindent\textbf{Obs.8. Exactness in Law Article Prediction.}
Table \ref{tab:more_results_laws} illustrates the model's capability in Law Article Prediction, a task demanding precise statutory grounding rather than generative flexibility. LegalGraphRAG achieves a superior overall accuracy of 47.9\%, establishing a substantial lead over both the strongest RAG baseline, HippoRAG2 (39.8\%), and the domain-specific state-of-the-art, ADAPT (41.3\%). Remarkably, our 8B-parameter framework even surpasses the massive DeepSeek-V3.1 (44.9\%), highlighting that our structured, evidence-based retrieval mechanism is more effective at pinpointing legal provisions than simply scaling model parameters or employing semantic retrieval.

\noindent\textbf{Obs.9. Precision in Term of Penalty Prediction.}
Table \ref{tab:more_results_terms} presents the results on the challenging term of penalty prediction task, which requires fine-grained quantitative reasoning rather than simple classification. LegalGraphRAG demonstrates a significant advantage in minimizing prediction error, consistently achieving the lowest Mean Absolute Error (MAE) across most subdomains compared to other RAG-based methods. For instance, in the Public Safety category, our model achieves an MAE of 20.9, outperforming RAPTOR (21.7) and HippoRAG2 (23.0). This indicates that while exact term matching remains difficult for all models, LegalGraphRAG’s evidence-based retrieval strategy effectively locates relevant sentencing guidelines and comparable precedents, thereby constraining the generation to a more precise and legally grounded time range.

\subsection{Hyper-parameter Sensitivity (Q5)}
\noindent To evaluate system stability, we investigated the sensitivity of the Researcher Agent to the retrieval parameter $k$, which governs the number of semantic concepts retrieved from the ontology graph $\mathcal{G}_{ont}$. We varied $k$ over the set $\{3, 4, 5, 6\}$. The upper bound is restricted to 6, as empirical evidence suggests that exceeding this threshold introduces excessive context noise, which overwhelms the model's effective window and degrades reasoning.

\begin{figure}[h]
    \centering
    \includegraphics[width=0.9\linewidth]{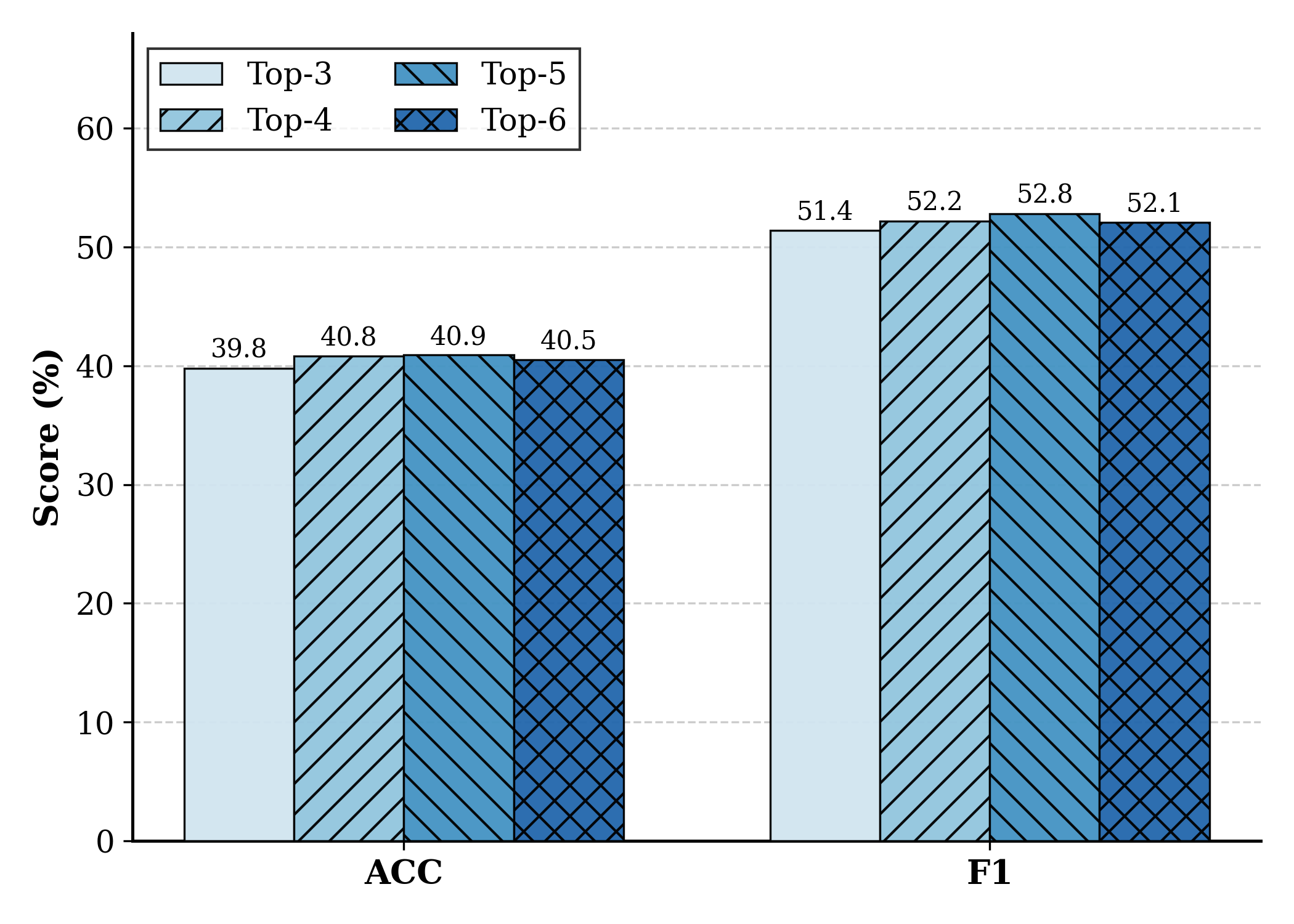}
    \caption{Impact of the retrieval parameter $k$ on charge prediction performance (CAIL dataset). The backbone model is Qwen3-8B.} 
    \label{fig:hyperparameter}
    \vspace{-2mm}
\end{figure}

\noindent\textbf{Obs.10. Robustness to Retrieval Hyperparameter Variations.} As illustrated in Figure~\ref{fig:hyperparameter}, LegalGraphRAG exhibits strong robustness to variations in $k$. Although performance peaks at $k=5$ (achieving 40.9 Accuracy and 52.8 F1), the variance across the tested range is marginal. This indicates that the Researcher Agent reliably captures essential semantic information without being hypersensitive to exact thresholding, provided $k$ remains within a reasonable bound.

\subsection{Latency and Token Cost (Q6)}
In this section, we shift our focus to the practical efficiency and computational overhead of the compared frameworks. Beyond accuracy, the latency and token consumption are crucial factors for real-world deployment. We provide a detailed breakdown and comparison of the computational costs for RAPTOR, HippoRAG2, and LegalGraphRAG across two primary phases: (i) the offline graph construction stage, reporting both the time and token cost required to build the knowledge base; and (ii) the online query-answering stage, reporting the time and token cost incurred during the retrieval and reasoning process for a given query.

\noindent\textbf{Obs.11. Trade-off between Efficiency and Interpretability.} Table \ref{tab:cost} presents the computational efficiency. LegalGraphRAG demonstrates superior offline efficiency with the lowest indexing time ($3687.49$s) and token consumption. However, during the online phase, it incurs higher latency ($46.1$s) and token usage. This overhead is a necessary trade-off for evidence-based reasoning. Unlike baseline GraphRAG approaches that often operate as opaque "black boxes", our method explicitly constructs credible reasoning chains to support its judgments. While generating such transparent evidence consumes more resources, it is indispensable for ensuring the trustworthiness and interpretability required in legal domains.

\section{Implementation Details}\label{sec:Implementation Details}
\subsection{Benchmark Dataset}\label{sec:Benchmark Dataset}
We evaluate LegalGraphRAG on two benchmark datasets, including CAIL2018\cite{xiao2018cail2018} and CMDL\cite{huang2024cmdl}.

\noindent\textbf{CAIL2018}\cite{xiao2018cail2018}: A large-scale Chinese legal dataset designed for the task of Legal Judgment Prediction (LJP), where models predict court outcomes based on factual case descriptions. It comprises over 2.6 million criminal cases published by the Supreme People’s Court of China, making it the largest publicly available dataset of its kind. Each case includes a detailed fact description along with structured judgment annotations, namely applicable law articles (183 categories), charges (202 categories), and prison terms. The dataset was created to address the lack of high-quality, large-scale resources in legal AI and to provide a realistic benchmark that reflects the complexity and imbalance inherent in real judicial data, where frequent charges dominate the case distribution. CAIL2018 has since become a foundational resource for evaluating and advancing automated legal judgment prediction systems.

\noindent\textbf{CMDL}\cite{huang2024cmdl}: A large-scale, real-world Chinese Multi-Defendant Legal Judgment Prediction dataset designed to address the under-explored challenge of predicting judicial outcomes in cases involving multiple defendants. It comprises 393,945 criminal cases with approximately 1.2 million defendants, covering 321 distinct charges and 275 legal articles. Notably, CMDL introduces case-level evaluation metrics that account for case complexity and varying numbers of defendants, offering a more holistic assessment of model performance in multi-defendant scenarios. For experimental feasibility, the subset \textbf{CMDL-small} is often utilized, as it preserves the data distribution while significantly reducing computational costs, making it suitable for preliminary benchmarking and model validation in resource-constrained research settings.

\noindent\textbf{Dataset Construction}\quad Due to considerations regarding the generation speed and token cost of the GraphRAG method, we constructed focused subsets from both the CAIL2018 and CMDL datasets using a uniform procedure aimed at controlling input length, balancing charge distribution, and elevating task complexity. The construction first filtered cases to retain only those with factual descriptions under 1,024 characters. To ensure broader coverage of under-represented charges and increase predictive difficulty, the sampling prioritized defendants whose charges included low-frequency offenses and deliberately retained a higher proportion of multi-charge cases. Consequently, the resulting subsets feature a more balanced charge distribution with elevated presence of rare charges and greater average case complexity compared to the original datasets. While this design provides a more challenging testbed for evaluating model performance on complex and low-frequency legal scenarios, it may also lead to lower reported performance for some methods relative to their results on the original, more naturalistic data distribution. Table \ref{tab:datasets} presents detailed statistics of the subsets.

\begin{table}
\centering
\footnotesize
\begin{tabular}{@{}lcc@{}}
\toprule
Dataset & CAIL & CMDL \\
\midrule
\# Case Num & 568 & 572  \\
\# Charges  & 168 & 239  \\
\# Average criminal per case  & 1.25 & 2.40  \\
\# Average defendant per case &  1.72 &  1.14 \\
\# Average length per case & 654.79 & 517.13  \\
\bottomrule
\end{tabular}
\caption{Basic statistics of the test datasets.}
\label{tab:datasets}
\end{table}
\subsection{Corpus}\label{sec:Corpus}
In this section, we provide detailed descriptions of corpus used in our experiments. To ensure comprehensive coverage of criminal statutes and charges, we construct this knowledge base by aggregating a subset of cases from multiple authoritative legal datasets: JuDGE\cite{su2025judge}, CAIL2018\cite{xiao2018cail2018}, CMDL\cite{huang2024cmdl}, and LeCaRDv2\cite{li2024lecardv2}. The construction follows a procedure similar to that used for the experimental subsets: we first filter cases by fact description length (under 1,024 characters) and apply sampling designed to balance the representation of different charges, thereby creating a broad and diverse collection of historical precedents and factual patterns. Furthermore, to ground the system in authoritative legal provisions, we incorporate the full text of the ``Criminal Law of the People's Republic of China'' along with its relevant judicial interpretations as a core statutory knowledge library. The combination of this curated historical case library and the official legal provisions library forms the complete corpus, enabling models to retrieve both experiential precedents and statutory knowledge during reasoning.Crucially, we have carefully verified that all cases in the corpus are distinct from those in the test subsets, ensuring no data leakage between the knowledge base and the evaluation benchmarks. The size and composition statistics of the final corpus are detailed in Table \ref{tab:case rag datasets} \& \ref{tab:knowledge rag datasets}.

\begin{table}
\centering
\footnotesize
\begin{tabular}{@{}lc@{}}
\toprule
Dataset & Case \\
\midrule
\# Num & 14049  \\
\# Charges*  & 818  \\
\# Average defendant per case &  3.39 \\
\# Average length per case & 399.73  \\
\bottomrule
\end{tabular}
\caption{Basic statistics of the cases in corpus. *The large number of ``Charges'' is due to inconsistencies in the descriptions of crimes across different datasets.}
\label{tab:case rag datasets}
\end{table}

\begin{table}
\centering
\footnotesize
\begin{tabular}{@{}lcc@{}}
\toprule
Dataset & Article &  Judicial interpretations \\
\midrule
\# Num & 452 & 656  \\
\# Average length & 128.54 & 243.95  \\
\bottomrule
\end{tabular}
\caption{Basic statistics of the legal knowledges in corpus.}
\label{tab:knowledge rag datasets}
\end{table}

\subsection{Evaluation Metrics}\label{sec:Evaluation Metrics}
To comprehensively evaluate the Legal Judgment Prediction (LJP) tasks, we employ specific metrics for different sub-tasks: Charge and Article Prediction are evaluated using Accuracy (ACC) and Micro-F1, Term of Penalty Prediction is assessed using ACC and Mean Absolute Error (MAE), and the Retrieval Quality of our RAG system is measured by Retrieval Effectiveness and Error Rate.

\noindent\textbf{Accuracy (ACC)} measures the exact match ratio. For classification tasks, it requires the predicted label set \(O_i\) to be identical to the ground truth \(O'_i\). For Term of Penalty, it measures the exact match of the predicted term. It is calculated as: 
$$
\text{ACC} = \frac{1}{N} \sum_{i=1}^N \mathbb{I}(O_i = O'_i)
$$
Where \(\mathbb{I}(\cdot)\) is the indicator function.

\noindent\textbf{Micro-F1} is used for multi-label classification to account for class imbalance. It is the harmonic mean of micro-averaged precision (\(P_{\text{micro}}\)) and recall (\(R_{\text{micro}}\)): 
$$
\text{Micro-F1} = \frac{2 \cdot P_{\text{micro}} \cdot R_{\text{micro}}}{P_{\text{micro}} + R_{\text{micro}}}
$$

\noindent\textbf{Mean Absolute Error (MAE)} reflects the deviation in the predicted term of penalty. Let \(T_i\) and \(T'_i\) denote the predicted and ground-truth prison terms (in months), respectively. MAE is defined as: 
$$
\text{MAE} = \frac{1}{N} \sum_{i=1}^N |T_i - T'_i|
$$

\noindent\textbf{Retrieval Effectiveness} measures how well the retrieved content aligns with the question's intent. Higher values indicate more focused and pertinent information. It is defined as:
$$
\text{Retrieval Effectiveness} = \frac{1}{|\mathcal{C}|} \sum_{c \in \mathcal{C}} \mathrm{R}(c, Q, \mathcal{E})
$$
where \(\mathcal{C}\) denotes the set of retrieved contexts, \(Q\) represents the question, \(\mathcal{E}\) denotes the set of evidence, and the operator \(\mathrm{R}(\cdot)\) determines the relevance of a context \(c\).

\noindent\textbf{Error Rate} quantifies the incompleteness of the retrieval process. Instead of measuring recall directly, we assess the proportion of reference claims not supported by the retrieved context:
\begin{equation}
    \text{Error Rate} = 1 - \left( \frac{1}{|\mathcal{R}|} \sum_{c \in \mathcal{R}} \mathbb{I}(\mathrm{S}(c, \mathcal{C})) \right)
\end{equation}

where \(\mathcal{R}\) is the set of reference claims, \(\mathrm{S}(\cdot)\) determines whether a claim \(c\) is supported by the retrieved context \(\mathcal{C}\), and $\mathbb{I}(\cdot)$ is the indicator function. A lower Error Rate indicates a more comprehensive evidence collection.

\subsection{Baseline Details}\label{sec:Baseline Details}
In this section, we provide detailed descriptions of each baseline used in our comparison, as detailed in figure~\ref{fig:rag_configs}.

\noindent\textbf{Naive RAG}\quad uses the standard RAG paradigm: a retriever model first retrieves relevant context from the corpus based on the given question, and then the question is concatenated with the retrieved context to form a query for the generation model to produce the final answer.

\noindent\textbf{G-retriever}\cite{he2024g}\quad introduces a retrieval augmented generation framework for textual graphs by formulating subgraph retrieval as a Prize-Collecting Steiner Tree optimization problem, enabling conversational question answering across diverse domains like scene understanding and knowledge graphs while mitigating LLM hallucinations and scaling to large graph sizes.

\noindent\textbf{LightRAG}\cite{guo2024lightrag}\quad introduces a graph-enhanced retrieval-augmented generation framework that integrates entity-relationship graphs into text indexing, combining low-level precise entity retrieval with high-level thematic discovery for efficient and adaptive knowledge integration.

\noindent\textbf{HippoRAG2}\cite{gutierrez2025rag}\quad builds on HippoRAG's Personalized PageRank framework by integrating dense-sparse coding for passages and phrases in the knowledge graph, enabling deeper contextualization and recognition memory for triple filtering. Enhances online retrieval with query-to-triple matching and optimized seed node weighting, outperforming standard RAG across factual, sense-making, and associative memory tasks.

\noindent\textbf{RAPTOR}\cite{sarthi2024raptor}\quad constructs a hierarchical tree by recursively clustering and summarizing embedded text chunks, enabling retrieval of information at multiple levels of abstraction to improve performance on long-document question-answering tasks.

\begin{figure}[htbp] 
\centering

\begin{tcolorbox}[
    colframe=black,        
    colback=gray!15,      
    coltitle=white,        
    fonttitle=\bfseries,   
    title=RAG Configuration
]
\begin{verbatim}
{
  embedding_model: bge-m3,
  retrieval_topk: 5,
  chunk_token_size: 1000,
  chunk_overlap_token_size: 200
}
\end{verbatim}
\end{tcolorbox}

\begin{tcolorbox}[
    colframe=black,        
    colback=gray!15,      
    coltitle=white,        
    fonttitle=\bfseries,   
    title=RAG Configuration
]
\begin{verbatim}
{
  embedding_model: bge-m3,
  retrieval_topk: 5,
  chunk_token_size: 1000,
  chunk_overlap_token_size: 200
}
\end{verbatim}
\end{tcolorbox}

\begin{tcolorbox}[
    colframe=black,        
    colback=gray!15,      
    coltitle=white,        
    fonttitle=\bfseries,   
    title=G-retriever Configuration
]
\begin{verbatim}
{
  embedding_model: bge-m3,
  retrieval_topk: 3,
  chunk_token_size: 1200,
  chunk_overlap_token_size: 100,
  entities_max_tokens: 2000,
  relationships_max_tokens: 2000
}
\end{verbatim}
\end{tcolorbox}

\begin{tcolorbox}[
    colframe=black,        
    colback=gray!15,      
    coltitle=white,        
    fonttitle=\bfseries,   
    title=LightRAG Configuration
]
\begin{verbatim}
{
  embedding_model: bge-m3,
  query_type: hybrid,
  chunk_token_size: 1200, 
  retrieval_topk: 20, 
  chunk_overlap_token_size: 100, 
  max_token_text_unit: 2000, 
  max_token_global_context: 2000, 
  max_token_local_context: 2000
}
\end{verbatim}
\end{tcolorbox}

\begin{tcolorbox}[
    colframe=black,        
    colback=gray!15,      
    coltitle=white,        
    fonttitle=\bfseries,   
    title=HippoRAG2 Configuration
]
\begin{verbatim}
{
  embedding_model: bge-m3,
  retrieval_top_k: 5, 
  linking_top_k: 5, 
  max_qa_steps: 3, 
  qa_top_k: 5,
  graph_type: facts_and_sim_passage
  _node_unidirectional
}
\end{verbatim}
\end{tcolorbox}

\end{figure}

\begin{figure}[htbp] 
\centering

\begin{tcolorbox}[
    colframe=black,        
    colback=gray!15,      
    coltitle=white,        
    fonttitle=\bfseries,   
    title=RAPTOR Configuration
]
\begin{verbatim}
{
  embedding_model: bge-m3,
  chunk_token_size: 1200, 
  chunk_overlap_token_size: 100, 
  num_layers: 5, 
  max_length_in_cluster: 3500, 
  threshold: 0.1, 
  cluster_metric: cosine, 
  threshold_cluster_num: 5000
}
\end{verbatim}
\end{tcolorbox}

\caption{Hyperparameter configurations for the baseline RAG models.}
\label{fig:rag_configs} 
\end{figure}

\noindent\textbf{Disc-LawLLM}\cite{yue2024lawllm}\quad is a retrieval-augmented large language model fine-tuned on Chinese judicial datasets using legal syllogism prompting to provide reasoning-capable legal services, including consultation, judgment prediction, and examination assistance. In this work, we use the officially open-sourced LawLLM-7B, which is fine-tuned from Qwen2.5-Instruct-7B.

\noindent\textbf{Legal $\Delta$}\cite{dai2025legal}\quad employs a reinforcement learning framework that enhances legal reasoning in LLMs by maximizing chain-of-thought guided information gain through dual-mode inputs and differential Q-value analysis. In this work, we use the officially open-sourced model, which is fine-tuned from Qwen2.5-Instruct-7B.

\noindent\textbf{ADAPT}\cite{deng2024enabling}\quad is a discriminative reasoning framework for LLMs in legal judgment prediction that emulates human judicial processes by asking to decompose case facts into key elements, discriminating among candidate charges for alignment, and predicting final judgments, further improved via multi-task fine-tuning with synthetic trajectories. In this work, we use the officially open-sourced model, which is fine-tuned from Qwen2-7B.

\subsection{LegalGraphRAG Setup}\label{sec:LegalGraphRAG Setup}
In the experimental setup of LegalGraphRAG, hyperparameters are configured to optimize retrieval precision. During graph construction, the Ontology Graph utilizes k-nearest neighbors (kNN) to select the top-3 case feature nodes based on cosine similarity for direct semantic matching.

For the Researcher agent, the evidence-based retrieval strategy operates with specific thresholds: the retrieval parameter is set to $k=5$. This parameter governs both the number of top-ranked semantic concepts retrieved from the ontology graph and the scope of community expansion, a value selected based on the sensitivity analysis to balance context coverage and noise control.

\begin{figure*}[t]
    \centering
    \includegraphics[width=0.95\linewidth]{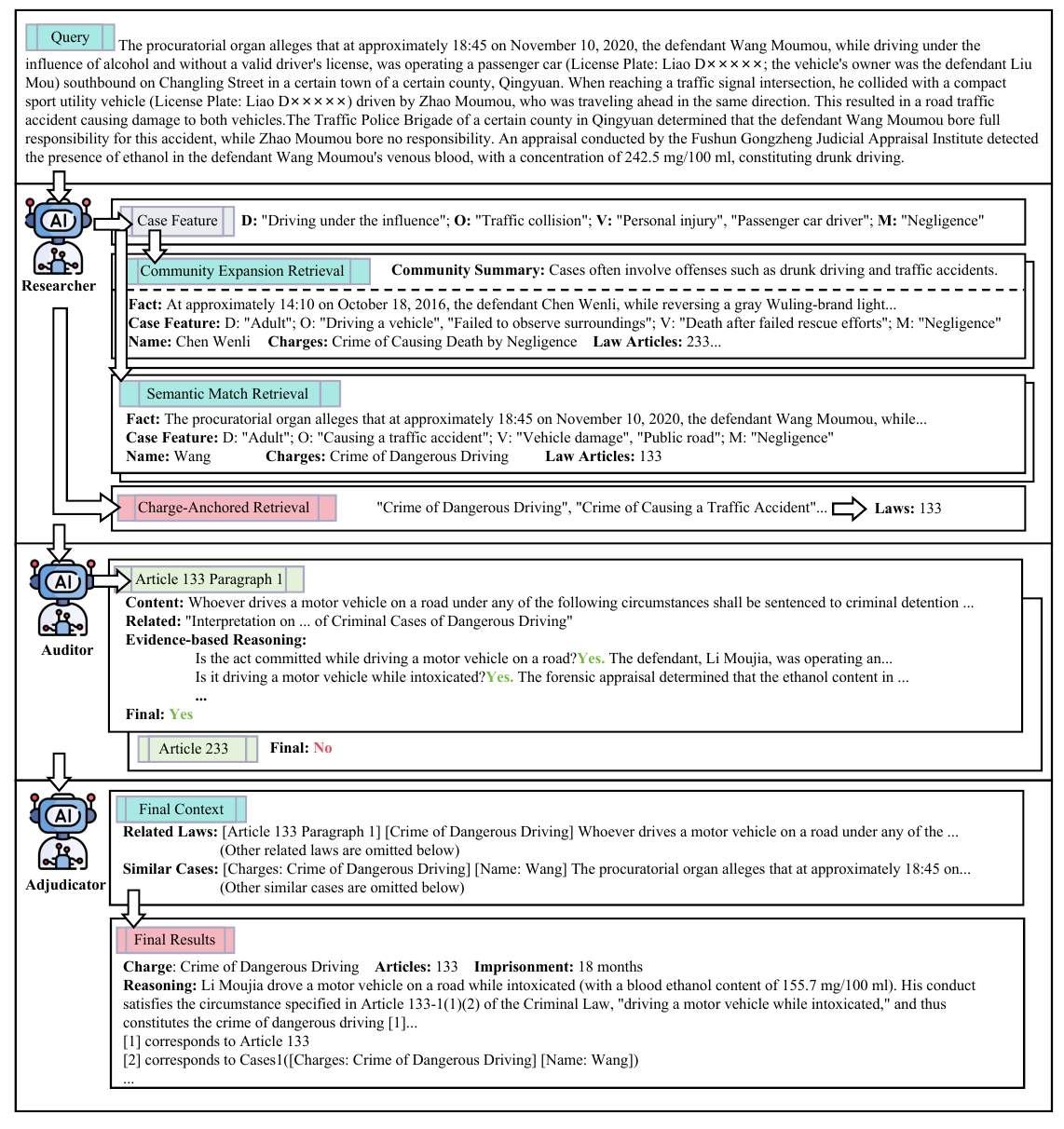}
    \caption{Qualitative analysis of a representative case regarding the crime of \textbf{Dangerous Driving}. The visualization highlights the retrieval of evidence related to specific statutory conditions.}
    \label{fig:case1_in_appendix}
    \vspace{-2mm}
\end{figure*}

\begin{figure*}[t]
    \centering
    \includegraphics[width=0.95\linewidth]{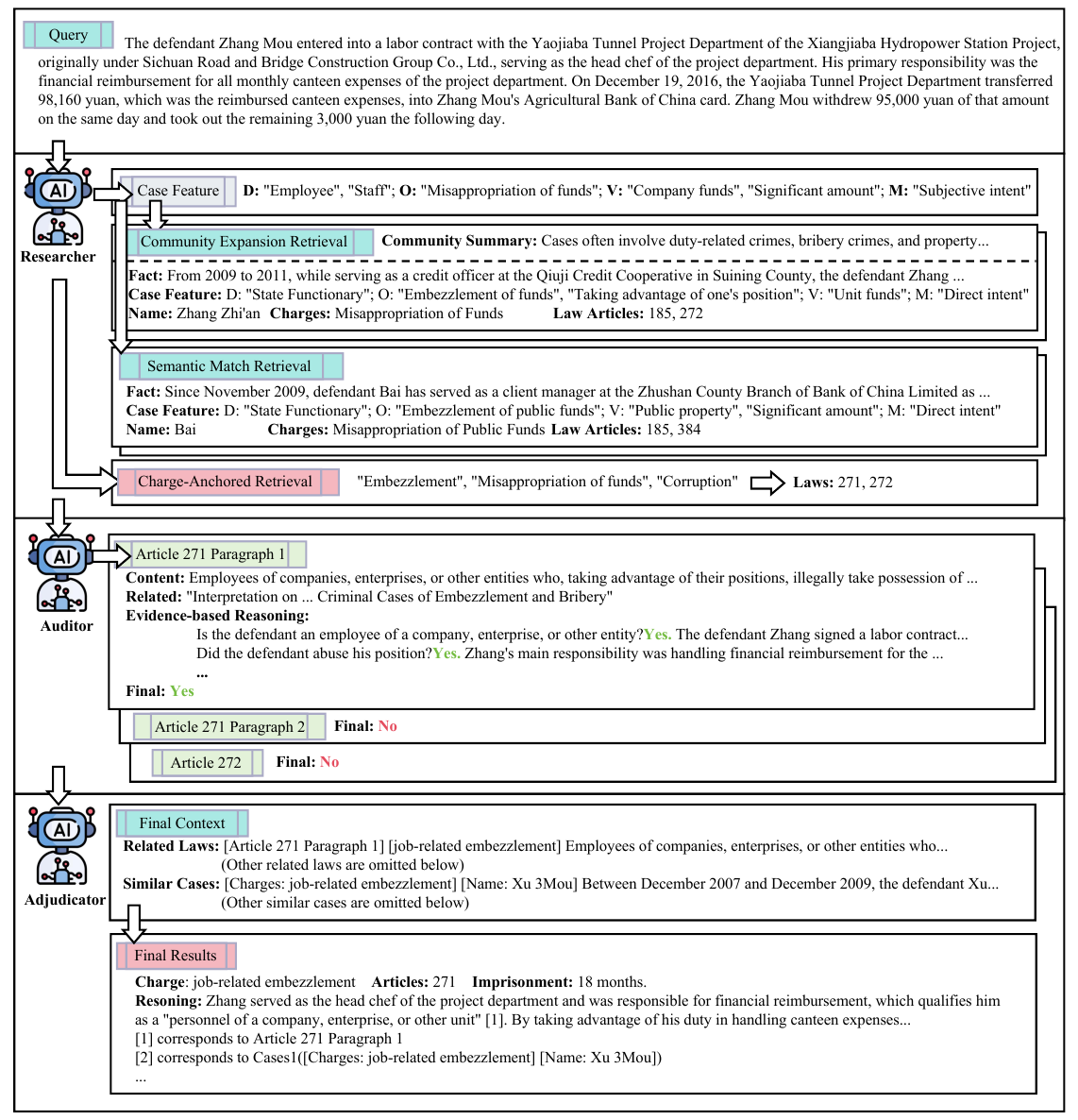}
    \caption{Qualitative analysis of a representative case regarding the crime of \textbf{Occupational Embezzlement}. The example demonstrates the model's reasoning in identifying the abuse of professional position.}
    \label{fig:case2_in_appendix}
    \vspace{-2mm}
\end{figure*}

\section{Extended Case Examples}\label{sec:Extended Case Examples}
In this section, we will walk through several cases to detail the retrieval and reasoning pipeline of LegalGraphRAG. Figure \ref{fig:case1_in_appendix} illustrates evidence retrieval for a Dangerous Driving case, while Figure \ref{fig:case2_in_appendix} demonstrates the legal reasoning process for an Occupational Embezzlement case.

\section{Prompt Set}
To facilitate reproducibility and provide transparency into the agent behaviors, we present the specific instruction sets designed for the \textbf{Researcher} (Figures~\ref{fig:prompt_keyword} and \ref{fig:prompt_prejudge}), \textbf{Auditor} (Figures~\ref{fig:prompt_auditor_item} and \ref{fig:prompt_auditor_final}), and \textbf{Adjudicator} (Figures~\ref{fig:prompt_sentencing} and \ref{fig:prompt_verdict}) agents. These prompts orchestrate the multi-stage reasoning process described in the main text.

\definecolor{promptGreen}{RGB}{120, 195, 120} 
\definecolor{promptBlue}{RGB}{110, 120, 190}  

\newtcolorbox{mypromptbox}[2][]{
    enhanced,                 
    drop shadow,              
    colframe=#2,              
    colback=white,            
    coltitle=white,           
    title=\textbf{\Large #1}, 
    fontupper=\ttfamily,      
    sharp corners,            
    boxrule=1.5mm,            
    left=3mm, right=3mm, top=2mm, bottom=2mm, 
    arc=0mm,
    outer arc=0mm
}

\begin{figure*}[!ht] 
\begin{mypromptbox}[Criminal Case Keyword Extraction]{promptGreen}
\textbf{Task Definitions:} Extract legal keywords from a criminal case description and classify them into four categories: \textit{Defendant Attributes}, \textit{Criminal Behaviors}, \textit{Victim Characteristics}, and \textit{Subjective Mental States}. The output must be a strictly valid JSON object without additional text.

\textbf{Keyword Definitions:}
\begin{itemize}[noitemsep, topsep=0pt, parsep=0pt, leftmargin=*]
    \item \textbf{Defendant Attributes:} Legal traits (e.g., age group, criminal history, occupation). Avoid specific names or numbers.
    \item \textbf{Criminal Behaviors:} Legal types of acts and significant methods. Exclude specific time/location details.
    \item \textbf{Victim Characteristics:} Nature of the property or location. Generalize specific amounts (e.g., ``large amount'').
    \item \textbf{Subjective Mental States:} Legal descriptions of intent and remorse.
\end{itemize}

\textbf{Output Example:}
\texttt{\{ \\
    ``Defendant\_Attribute'': [``Adult'', ``Prior Criminal Record''], \\
    ``Criminal\_Behaviors'': [``Theft'', ``Burglary''], \\
    ``Victim\_Characteristics'': [``Private Residence'', ``Large Amount''], \\
    ``Subjective\_Mental\_States'': [``Direct Intent'', ``Voluntary Surrender''] \\
\}}
\end{mypromptbox}
\caption{Prompt for the Researcher agent to extract and classify legal keywords from case descriptions.}
\label{fig:prompt_keyword}

\vspace{0.3cm} 

\begin{mypromptbox}[Charge Pre-judge(for Charge-Anchored Retrieval)]{promptBlue}
\textbf{Task Definitions:} Act as a criminal law expert to analyze the provided case (\texttt{\{case\_text\}}).
\begin{itemize}[noitemsep, topsep=0pt, parsep=0pt, leftmargin=*]
    \item Output reasonably possible charges (confidence $>30\%$) sorted by probability (descending).
    \item If a dominant charge exists (confidence $>70\%$), prioritize it; if it is the only certain charge, output it exclusively.
    \item Exclude charges with probability $<10\%$.
\end{itemize}

\vspace{0.5em} 

\textbf{Format Definitions:}
\begin{itemize}[noitemsep, topsep=0pt, parsep=0pt, leftmargin=*]
    \item Output strictly as a Python list: \texttt{['Charge 1', 'Charge 2', ...]}.
    \item The output must start with \texttt{[}.
    \item Return an empty list \texttt{[]} if no charge matches.
    \item \textbf{No} additional explanations or text allowed.
\end{itemize}
\end{mypromptbox}
\caption{Prompt for the Researcher agent to pre-judge potential charges for charge-anchored retrieval.}
\label{fig:prompt_prejudge}
\end{figure*}

\begin{figure*}[!ht] 
\begin{mypromptbox}[Auditor Checklist(item)]{promptGreen}
\textbf{Task Definitions:} Act as a legal AI assistant to assess if the case facts strictly satisfy a specific constituent element of the law.
\begin{itemize}[noitemsep, topsep=0pt, parsep=0pt, leftmargin=*]
    \item Analyze the \texttt{law\_item} and \texttt{case} facts.
    \item Focus exclusively on the target \texttt{element} (e.g., ``intent''), using \texttt{related} materials (if provided) for interpretation.
    \item determine applicability based on facts and logic.
\end{itemize}

\vspace{0.5em}

\textbf{I/O Specifications:}
\begin{itemize}[noitemsep, topsep=0pt, parsep=0pt, leftmargin=*]
    \item \textbf{Input:} \texttt{law\_item}, \texttt{related} (supplementary materials), \texttt{element}, \texttt{case}.
    \item \textbf{Output:} Provide reasoning first, then enclose the final result strictly within tags: \texttt{<answer>true</answer>} or \texttt{<answer>false</answer>}.
\end{itemize}

\vspace{0.5em}
\textbf{Template:} \\
law: \texttt{\{law\_item\}}, related: \texttt{\{related\}} \\
element: \texttt{\{element\}}, case: \texttt{\{case\}}
\end{mypromptbox}
\caption{Prompt for the Auditor agent to verify if case facts satisfy specific constituent elements of the law.}
\label{fig:prompt_auditor_item}

\vspace{0.3cm} 

\begin{mypromptbox}[Auditor Checklist(final)]{promptBlue}
\textbf{Task Definitions:} Act as a legal analysis assistant to determine if the provided law article applies to the specific case (i.e., verify violation or crime).
\begin{itemize}[noitemsep, topsep=0pt, parsep=0pt, leftmargin=*]
    \item Identify all relevant constituent elements from the \texttt{law} text.
    \item Verify critical elements independently; note that the provided \texttt{true\_list} and \texttt{false\_list} may be incomplete.
\end{itemize}

\vspace{0.5em}

\textbf{I/O Specifications:}
\begin{itemize}[noitemsep, topsep=0pt, parsep=0pt, leftmargin=*]
    \item \textbf{Input Variables:} \texttt{case}, \texttt{law}, \texttt{true\_list} (proven elements), \texttt{false\_list} (disproven elements).
    \item \textbf{Output:} Provide reasoning first, then enclose the final result strictly within tags: \texttt{<answer>true</answer>} or \texttt{<answer>false</answer>}.
\end{itemize}

\vspace{0.5em}
\textbf{Template:} \\
case: \texttt{\{case\}}, law: \texttt{\{law\}} \\
true\_list: \texttt{\{true\_list\}}, false\_list: \texttt{\{false\_list\}}
\end{mypromptbox}
\caption{Prompt for the Auditor agent to assess the overall applicability of a law based on verified elements.}
\label{fig:prompt_auditor_final}
\vspace{-5mm}
\end{figure*}

\begin{figure*}[p] 
\begin{mypromptbox}[Charge \& Sentencing (JSON)]{promptGreen}
\textbf{Task Definitions:} Act as a legal expert to adjudge the defendant based on candidate charges.
\begin{itemize}[noitemsep, topsep=0pt, parsep=0pt, leftmargin=*]
    \item \textbf{1. Final Charge Application:} For concurrence, apply the ``heavier penalty'' rule; for multiple acts, apply combined punishment.
    \item \textbf{2. Sentencing:} Predict the specific law article and a reasonable sentencing range based on facts and judicial practice.
\end{itemize}

\vspace{0.5em}

\textbf{Format Definitions:}
\begin{itemize}[noitemsep, topsep=0pt, parsep=0pt, leftmargin=*]
    \item \begin{verbatim}
{
  charge_name: [Charge A, ...],
  law_article: [Art. X, ...],
  term_of_imprisonment: {
    death_penalty: boolean,
    imprisonment: integer (months),
    life_imprisonment: boolean
  }
}
    \end{verbatim}
\end{itemize}
\end{mypromptbox}
\caption{Prompt for the Adjudicator agent to generate structured sentencing predictions and apply legal rules.}
\label{fig:prompt_sentencing}

\vspace{0.3cm} 

\begin{mypromptbox}[Legal Reasoning \& Verdict]{promptBlue}
\textbf{Task Definitions:} Act as a legal consultant to analyze the case using the provided \texttt{Context} documents.
\begin{itemize}[noitemsep, topsep=0pt, parsep=0pt, leftmargin=*]
    \item \textbf{Step 1: Fact \& Act Analysis:} Analyze how many independent criminal acts exist. Explicitly cite the supporting evidence from the context using \texttt{[1][2]...}.
    \item \textbf{Step 2: Law Application:} Resolve any legal concurrence (e.g., Imaginative Concurrence vs. Combined Punishment). Explain why specific articles apply over others.
    \item \textbf{Step 3: Sentencing Prediction:} comprehensive assessment of sentencing based on statutory rules.
\end{itemize}

\vspace{0.5em}

\textbf{Output Format:}
\begin{itemize}[noitemsep, topsep=0pt, parsep=0pt, leftmargin=*]
    \item \textbf{Structure:} Output in two clear sections: \texttt{Legal Analysis} (reasoning with citations) and \texttt{Final Verdict} (conclusion).
    \item \textbf{Requirement:} You \textbf{must} mark the source of your facts or laws using brackets like \texttt{[1]}.
\end{itemize}

\vspace{0.5em}
\textbf{Template:} \\
Context: \texttt{\{context\_list\}}, Case: \texttt{\{case\_description\}}
\end{mypromptbox}
\caption{Prompt for the Adjudicator agent to synthesize legal reasoning and output the final verdict.}
\label{fig:prompt_verdict}

\end{figure*}

\section{Related Works}
\subsection{Legal Large Language Models}
The rapid evolution of LLMs has catalyzed the development of domain-specific models tailored for the legal sphere. For Chinese law, \textbf{ChatLaw}~\cite{cui2023chatlaw}, \textbf{DISC-LawLLM}~\cite{yue2023disc}, and \textbf{InternLM2Law}~\cite{fei2025internlm} leverage extensive legal corpora, including judicial interpretations and statutes, to handle diverse legal tasks. Other notable models like \textbf{LawGPT}~\cite{zhou2024lawgpt} and \textbf{Fuzi-Mingcha}~\cite{sdu2023fuzimingcha} integrate unsupervised legal texts with supervised fine-tuning to enhance domain understanding. Beyond Chinese, \textbf{SaulLM}~\cite{colombo2024saullm} focuses on English legal texts based on the Mixtral architecture, while \textbf{LawLLM}~\cite{shu2024lawllm} addresses US legal tasks such as similar case retrieval. Additionally, specialized models like \textbf{InLegalLLaMA}~\cite{ghosh2024inlegalllama} target Indian and French legal domains respectively. These models provide crucial baselines for downstream tasks but often lack the specific reasoning architecture required for complex judgment prediction.

\subsection{Legal judgment prediction}
Legal judgment prediction (LJP) has experienced significant development and become an increasingly crucial NLP task. Earlier research~\cite{segal1984predicting} relied on artificially designed features, and traditional machine learning methods~\cite{sulea2017exploring} were applied to predict legal judgments. Recent advances in deep learning~\cite{xu2020distinguish,han2023case} have motivated researchers to leverage neural networks for automated text representation learning. 
Recently, LLMs have further promoted the progress of LJP~\cite{deng2024learning}. Several studies~\cite{wu2023precedent,peng2024athena} employ Retrieval-Augmented Generation (RAG) to enhance LLMs by incorporating external legal knowledge. 
To refine decision-making, recent works have introduced structured reasoning frameworks that systematically decompose case facts to distinguish confusing charges~\cite{jiang2023legal,deng2024enabling,wang2024legalreasoner}. Furthermore, multi-agent simulation frameworks have been explored to improve performance by simulating court debates and analyzing cases from diverse perspectives~\cite{he2024agentscourt}.
However, existing LLM-based methods still struggle to utilize comprehensive legal knowledge~\cite{fei2024lawbench} effectively. In this context, we make full use of external knowledge and precedents within a unified framework.

\subsection{Reasoning skills in legal domain}
Recent work has improved LLMs’ reasoning through better prompting techniques~\cite{sahoo2024systematic}. Chain-of-thought (CoT)~\cite{wei2022chain} prompting can explicitly guide LLMs to reason step by step. In the legal domain, researchers have adapted CoT to legal-specific frameworks. For instance, Yu et al.~\cite{yu2022legalpromptingteachinglanguage} demonstrated that incorporating the IRAC (Issue, Rule, Application, Conclusion) framework significantly enhances reasoning capabilities.
LoT~\cite{jiang2023legal} proposed legal syllogism reasoning to improve performance on LJP tasks, and ADAPT~\cite{deng2024enabling} established a workflow enabling discriminative reasoning. 
Moreover, approaches like MALR~\cite{yuan-etal-2024-large} utilize parameter-free learning to decompose complex legal tasks, while CaseGPT~\cite{yang2024casegptcasereasoningframework} combines LLMs with RAG to support semi-structured reasoning and legal argumentation~\cite{westermann2024dallma}.
Additionally, GLARE~\cite{yang2025glare} leverages an agentic framework and web data for legal reasoning. However, these approaches primarily rely on intrinsic capabilities or noisy external data, which constraints reasoning depth~\cite{zhang2024should,ke2025survey}. Therefore, we propose an agentic framework to dynamically acquire key legal knowledge, enhancing both breadth and depth.

\subsection{RAG in legal domain}
In the legal domain, recent studies have adapted RAG and graph-based solutions for specific tasks, particularly Legal Question Answering (LQA) and retrieval pipelines.
For instance, recent works optimize LLM outputs by incorporating external case-based information~\cite{wiratunga2024cbr,louis2024interpretable} or utilizing adapt-retrieve-revise pipelines~\cite{wan2024reformulating} to combine continual training with evidence revision.
Dedicated benchmarks such as LegalBench-RAG show that the retrieval stage remains a bottleneck~\cite{pipitone2024legalbench}. Works enriching retrieval with structural information (e.g., graphs of articles) demonstrate gains in tasks like statutory article retrieval~\cite{louis2023finding,hei2024heterogeneous,ho2025incorporating}. Recently, the SAT-Graph RAG framework~\cite{de2025graph} was proposed to model the hierarchical structure of legal norms. However, its sophisticated ontology-driven approach requires heavily structured input data, limiting its applicability to less curated corpora.

\section*{The Usage of LLMs}
In this paper, LLMs were used only to polish the writing and correct grammatical errors for clarity.
In the preparation of this manuscript, we utilized Large Language Models (LLMs) to assist with the writing process. Specifically, the model was used to refine the English text, including correcting grammatical errors and improving sentence clarity. Additionally, LLMs assisted in the initial formatting of several tables. The authors reviewed all model suggestions and retain full responsibility for the scientific accuracy and integrity of the final content.

\end{document}